\pdfoutput=1

\documentclass[11pt]{article}

\usepackage[final]{acl}  

\usepackage{times}
\usepackage{latexsym}

\usepackage[T1]{fontenc}

\usepackage[utf8]{inputenc}

\usepackage{microtype}

\usepackage{inconsolata}

\usepackage{graphicx}
\usepackage{float}

\definecolor{myyellow}{HTML}{FFC000}

\usepackage{xspace}
\usepackage{soul}
\usepackage{url}
\usepackage{graphicx}
\usepackage{amsmath}
\usepackage{amsfonts}
\usepackage{multirow}
\usepackage{amsthm}
\usepackage{booktabs}
\usepackage{epstopdf}
\usepackage{xcolor}
\definecolor{lavenderblue}{HTML}{7030A0}

\usepackage{bbm}
\usepackage{dsfont}
\usepackage{multirow, booktabs, tabularx}

\definecolor{lightblue}{RGB}{224,236,255}
\definecolor{graycell}{RGB}{238,238,238}

\usepackage{arydshln}

\usepackage{float} 
\usepackage{stfloats}   
\usepackage{setspace}
\usepackage{braket}
\usepackage{tablefootnote}
\usepackage[normalem]{ulem}
\usepackage{hyperref}
\usepackage{cleveref}
\usepackage{url}
\usepackage{makecell}
\usepackage{wrapfig}
\usepackage{sansmath}
\usepackage{float}
\usepackage{pifont}
\useunder{\uline}{\ul}{}
\usepackage{tcolorbox}
\usepackage{fancyvrb}
\usepackage{listings}
\usepackage[textsize=tiny]{todonotes}
\usepackage[toc,page,header]{appendix}
\usepackage{minitoc}

\newcommand{\ourmethod}{\texttt{MedEditBench}\xspace}

\graphicspath{{pics/}}
\newcommand{\ouredit}{SGR-Edit\xspace}

\usepackage{fancyhdr}
\title{Beyond Memorization: A Rigorous Evaluation Framework for \\ Medical Knowledge Editing}

\author{
  \textbf{Shigeng Chen}\textsuperscript{1}\thanks{~~Equal contribution.},~
  \textbf{Linhao Luo}\textsuperscript{2}\footnotemark[1],~
  \textbf{Zhangchi Qiu}\textsuperscript{1}\\[5pt] 
  \textbf{Yanan Cao}\textsuperscript{3},~
  \textbf{Carl Yang}\textsuperscript{4},~
  \textbf{Shirui Pan}\textsuperscript{1}\thanks{~~Corresponding author.}
  \\
  \textsuperscript{1}School of Information and Communication Technology, Griffith University\\
  \textsuperscript{2}Department of Data Science and AI, Monash University\\
  \textsuperscript{3}Institute of Information Engineering, Chinese Academy of Sciences\\
  \textsuperscript{4}Department of Computer Science, Emory University \\
  \texttt{shigeng.chen@griffithuni.edu.au}, \texttt{s.pan@griffith.edu.au}
}

\begin{document}
\maketitle




\begin{abstract}
Knowledge editing (KE) has recently emerged as a promising technique to update specific facts in large language models (LLMs) without full retraining. While existing KE methods show promising results on general-domain benchmarks, their effectiveness in the medical domain remains largely unexplored. Medical knowledge editing poses unique challenges, requiring models not only to memorize new facts but also to internalize and generalize them for reliable and interpretable clinical decision-making. In this work, we propose \ourmethod, a rigorous evaluation framework for assessing medical knowledge editing. Our preliminary results reveal that current KE paradigm, which directly edits simple answers to the LLMs, often leads to superficial updates with poor generalization. To address this, we introduce Self-Generated Rationale Editing (SGR-Edit), which leverages model-generated rationales as editing targets, enabling deeper knowledge integration. Extensive experiments across diverse LLMs and KE methods demonstrate that SGR-Edit consistently improves editing efficacy and generalization. Furthermore, we examine the impact of sequential edits on in-domain medical knowledge, external-domain knowledge, as well as general model capabilities, offering practical insights for deploying KE in real-world medical applications. Codes and datasets are available at \url{https://github.com/Aries-chen/MedEditBench}.


\end{abstract}
\section{Introduction}\label{sec:intro}

\sethlcolor{yellow} 

Large language models (LLMs) encapsulate extensive knowledge during training on large-scale corpora~\citep{store1,store2,store3}. Nonetheless, their knowledge remains static after training, resulting in factual inconsistencies and hallucinations in tasks that require up-to-date information or domain-specific expertise beyond their pre-trained knowledge \citep{hu2023large,huang2025survey}.
Knowledge editing (KE) has emerged as an approach that efficiently updates new knowledge in LLMs, allowing them to constantly adapt to the ever-evolving world without requiring full retraining \citep{rome,memit,alphaedit}.

Despite rapid advancements, the effectiveness of current KE methods has not been thoroughly assessed in realistic domain-specific scenarios. Most existing benchmarks only evaluate on general domains, such as WikiData \citep{rome} and counterfactual datasets \citep{zsre}, which do not reflect the complexity and diversity of real-world applications, particularly in specialized domains like medicine. 

\begin{figure}[t]
\centering
\includegraphics[width=1\columnwidth]{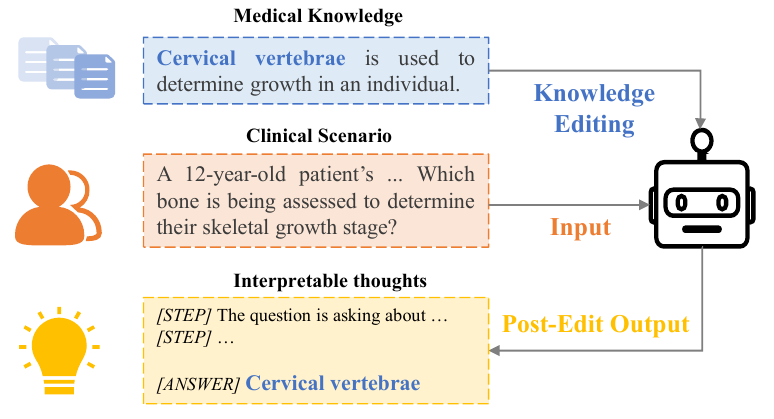}
\caption{Illustration of medical knowledge editing.}
\label{fig:intro}
\end{figure}

Medical knowledge editing is particularly challenging due to its intricacy and specificity, which requires LLMs to not only memorize updated knowledge \citep{gap1} but also comprehend underlying concepts and generalize to unseen scenarios \citep{gap2}, as shown in \Cref{fig:intro}. Crucially, the necessity for transparent and trustworthy medical decisions requires LLMs to also explain their reasoning to improve trust and interoperability \citep{exp1,exp2}.
Therefore, these challenges necessitate a more rigorous evaluation framework that accurately reflects the genuine performance of KE methods in the medical domain, ensuring their reliability and applicability in real-world applications \citep{app1,app2}.

In this work, we create a stringent medical knowledge editing benchmark based on two professional medical entrance examinations \citep{medexqa,medmcqa}. Specifically, we carefully curate a subset of QA pairs that current LLMs fail to answer correctly to serve as the target set for knowledge editing. 
To evaluate the generalizability and robustness of KE methods, we extend the benchmark with \emph{scenario-based questions} that require LLMs to apply the injected knowledge to address new clinical scenarios, as well as preservation questions that require LLMs to retain their previously factual knowledge.


Our findings show that existing KE methods fall short in complex medical settings: edited LLMs typically achieve less than 50\% efficacy, struggle to generalize knowledge to new scenarios, and often corrupt previously stored facts.
We attribute this pitfall to the existing editing paradigm that edits the LLMs through a simple answer, resulting in surface-level memorization rather than deep understanding, thereby impairing generalization and robustness.

Inspired by research on belief revision in logic and philosophy, where an intelligent agent updates its beliefs by incorporating new information while maintaining logical consistency and minimizing unnecessary changes \citep{logic,logic2}, we argue that knowledge editing should not merely overwrite outdated facts, but instead be guided by rational principles of consistency and justification. 
This perspective suggests that, much like human belief revision, effective editing requires encoding not only the updated fact but also the explanatory reasons that support it.
Building on this insight, we propose an editing paradigm called \textbf{S}elf-\textbf{G}enerated \textbf{R}ationale \textbf{Edit}ing (\ouredit), where LLMs are prompted to generate explanatory rationales~\citep{wei2022chain} grounded in reference texts. These self-generated rationales serve as the target knowledge that can be seamlessly integrated into arbitrary KE methods.
Experimental results demonstrate that SGR-Edit improves knowledge editing performance by up to 11.3\% across various LLMs and diverse KE methods, enabling edited LLMs to better internalize new medical knowledge and generalize to unfamiliar scenarios.
Our primary contributions are fourfold:
\begin{enumerate}
    \item \textbf{The \ourmethod Framework:} We introduce \ourmethod, a rigorous evaluation framework for medical knowledge editing. We employ a multi-faceted metric suite to comprehensively assess state-of-the-art KE methods in complex medical scenarios.
    
    \item \textbf{Pinpointing Paradigmatic Limitations:} We conduct extensive evaluations revealing that the conventional KE paradigm, which relies on editing simple final answers, primarily results in superficial memorization rather than genuine knowledge inter\-nal\-i\-za\-tion. This severely limits the generalization capability required for real-world practice.
    
    \item \textbf{Self-Generated Rationale Editing (SGR-Edit):} We propose a novel editing paradigm, SGR-Edit, utilizing the LLM's self-generated, evidence-based reasoning chain as the editing target. It significantly boosts editing performance across diverse LLMs and KE methods.
    
    \item \textbf{In-depth Analysis and Guidance:} We provide the first systematic analysis of sequential editing effects on internal/external knowledge and general capabilities. Our findings offer critical insights for deploying KE in safety-critical domains, guiding future research toward reliable and interpretable LLMs.
\end{enumerate}

\section{Related Work}
\label{sec:related}

\subsection{Knowledge Editing Methods}
Current KE methods can be grouped into three main categories.
The first, \textbf{Fine-Tuning–Based Editing}, injects new knowledge by updating a large number of model parameters, often through parameter-efficient training techniques \citep{ftl,ke1,lora}. The second category, \textbf{Parameter-Modifying Editing}, focuses on altering specific parameters to minimize interference with unrelated knowledge. This group is further divided into: 1) \emph{Meta-Learning} strategies, which train a hypernetwork to predict gradient updates for knowledge insertion \citep{ke,mend,malmen}; and 2) the \emph{Locate-then-Edit} approach, which directly identifies and rewrites factual weights in specific layers \citep{rome,memit}, and has recently evolved variants optimized for sequential editing \citep{alphaedit,anyedit}. Finally, \textbf{Parameter-Preserving Editing} maintains the base model's integrity by either augmenting it with external components \citep{grace,serac,wise} or using retrieval during inference \citep{ike,retrieval1,retrieval2,retrieval3} to store new facts.

\paragraph{Scope of This Work.} 
While retrieval-based approaches~\citep{retrieval1,retrieval2,retrieval3} have shown promise by augmenting models with external context, they do not alter the model's intrinsic knowledge. 
In this work, we distinguish ourselves by strictly focusing on \textit{model editing} (parameter updates), aiming to evaluate how well LLMs can \textit{intrinsically} internalize and generalize medical knowledge without relying on external retrieval modules.

\subsection{Knowledge Editing Benchmarks}

Current evaluations of KE methods primarily focused on general-domain benchmarks, such as WikiData \citep{rome} and counterfactual datasets \citep{zsre}, which aimed to update LLMs with new knowledge that contradicted common facts. 
However, recent studies have questioned the real-world applicability of these methods and proposed new, more rigorous benchmarks for fair evaluations. 
Specifically, \citet{ripple} measured ``ripple effects'' on related facts and revealed that most KE methods often fail to propagate consistent changes beyond the target triple. \citet{can} reported that prior benchmarks do not strictly confirm LLMs having hallucinated answers before editing, which masks their true editing performance. \citet{robustness} investigated editing consistency under prompt rephrasing and realistic contexts, finding that current KE methods exhibit lower generalization, and popular facts are the hardest to edit. 
Furthermore, \citet{Navigating} examined sequential editing and showed that KE methods experience performance drops after sequential edits, while \citet{mirage} critiqued common evaluation practices, revealing that KE methods catastrophically fail on realistic QA tasks. 

In contrast to previous evaluations focused on simplified general-domain settings \citep{robustness,Navigating}, we present the first medical knowledge editing evaluation framework \ourmethod, which not only rigorously measures the effective adoption of injected medical knowledge in new clinical scenarios without degrading previous knowledge but also systematically assesses the effects of sequential edits across internal knowledge, external knowledge, and model general abilities.
Furthermore, distinct from recent work focusing on long-tail knowledge \cite{long-tail}, our \ourmethod framework prioritizes the rigorous evaluation of reasoning stability and sequential editing robustness.


\section{Task Formulation}\label{sec:task}

The goal of knowledge editing is to modify specific knowledge $k$ in LLMs without retraining the entire model \citep{ke1}. This modification aims to improve performance on tasks related to that knowledge, represented by a set of queries and answers: $\mathcal{Q}^k=\{(q_i, a_i)\}$.
Let $\theta$ denote the original model. Given a query $q$ and the target knowledge $k$, the knowledge editing method $F$ can be expressed as:
\begin{equation}
    \theta' = F(\theta,q,k), \label{eq0}
\end{equation}
where $\theta'$ represents the edited model expected to provide the desired answer $a=\theta'(q)$ for the knowledge-related query $q\in\mathcal{Q}^k$.

\section{Evaluation Framework}\label{eval}

\begin{figure*}[!htb]
\centering
\includegraphics[width=1.0\textwidth]{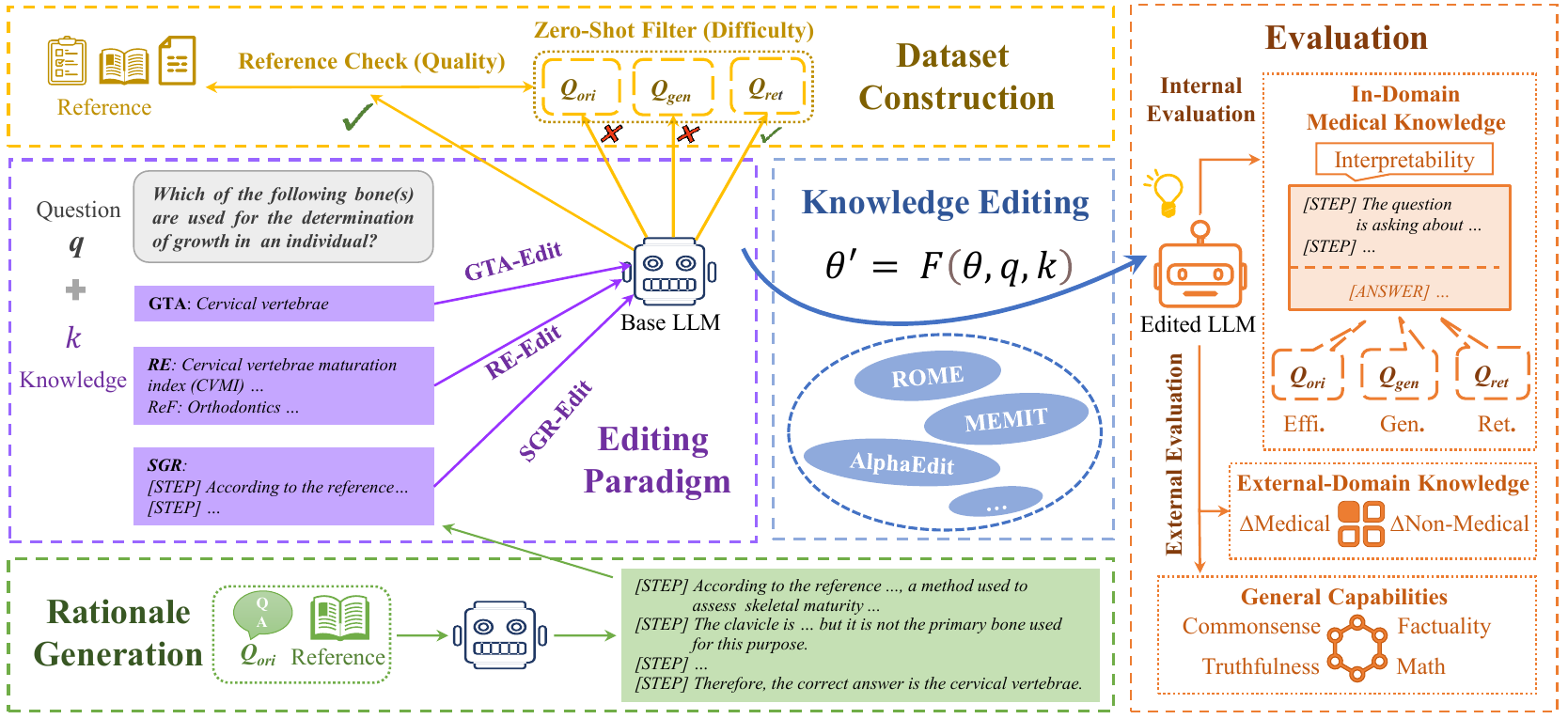}
  \caption{Overview of proposed medical knowledge editing evaluation framework (\ourmethod). Datasets are first constructed for medical knowledge editing (\textcolor{myyellow}{yellow rectangle}). 
  Three editing paradigms (GTA-Edit, RE-Edit, and SGR-Edit) are applied to evaluate the effectiveness of different target knowledge for medical knowledge editing (\textcolor{lavenderblue}{purple rectangle}). In particular, the proposed SGR-Edit generates a rationale by prompting the LLM itself, given a QA and a reference (\textcolor{green}{green rectangle}). Then, the knowledge is used to edit the base LLM using editing methods like ROME (\textcolor{blue}{blue rectangle}).
  In the final evaluation stage (\textcolor{orange}{orange rectangle}), the edited LLM is evaluated using \emph {efficacy}, \emph{generalization}, and \emph{retention}, as well as interpretability of the generated rationales. 
To examine how sequential medical edits affect the edited LLM’s broader knowledge and general abilities, we complement \ourmethod with external evaluations, including multi-domain knowledge (MMLU), and commonsense reasoning (HellaSwag), factual recall (TriviaQA), truthfulness (TruthfulQA), and mathematical reasoning (GSM8K).
}
\label{fig-overview}
\end{figure*}

In this section, we describe the proposed medical knowledge editing evaluation framework (\ourmethod), as illustrated in Figure \ref{fig-overview}.

\subsection{Medical Editing Benchmark Construction}

Due to the lack of existing benchmarks for medical knowledge editing, we construct two datasets: MedExQA\textsubscript{edit} and MedMCQA\textsubscript{edit} by extending two real-world medical QA datasets \citep{medexqa,medmcqa}. Each QA dataset is originally provided with a set of questions $q$, answers $a$, and relevant references $c$ in the format of human‐written explanations or textbook references.
To ensure reliable evaluations of a target LLM, we filter out medical questions that the LLM can answer correctly before editing. In this way, our assessments could reflect genuine improvements from knowledge editing methods in the rest set $\mathcal{Q}_{\mathrm{ori}}$. 
Thus, the accuracy on $\mathcal{Q}_{\mathrm{ori}}$ indicates the \textit{efficacy} of editing methods in injecting new knowledge.
To evaluate \textit{generalization} and \textit{retention}, we construct two auxiliary question sets: 
$\mathcal{Q}_{\mathrm{gen}}$, which extends original facts into new clinical scenarios, 
and $\mathcal{Q}_{\mathrm{ret}}$, which introduces off-topic questions to assess the preservation of unaffected knowledge. 
Examples of $\mathcal{Q}_{\mathrm{ori}}$, $\mathcal{Q}_{\mathrm{gen}}$, and $\mathcal{Q}_{\mathrm{ret}}$ are shown in \Cref{fig:QAs}.

\paragraph{Data Principles}
 
As shown in the top gray rectangle of \Cref{fig-overview}, our benchmarks are built following two key data principles: 
\textbf{Quality.} We filter each QA pair using \textit{Reference Check} to verify whether the provided context $c$ logically entails the ground‐truth answer $a$ for the question $q$.
We prompt the LLM with $(c,q)$ and retain the sample only if the predicted answer $\hat{a}$ equals $a$.
\textbf{Difficulty.} To strictly evaluate the efficacy and generalization of knowledge editing, we ensure the pre-edit performance of the LLM is 0\% on $\mathcal{Q}_{\mathrm{ori}}$ and $\mathcal{Q}_{\mathrm{gen}}$ using \textit{Zero-Shot Filter} (i.e., LLM's explicit incorrectness serves as direct signal for selecting examples to be targeted for editing and generalization evaluation), while 100\% on $\mathcal{Q}_{\mathrm{ret}}$ to ensure the retention of original knowledge.
Prompts and details for data construction and verification are provided in Appendix \ref{workflow}.

\subsection{Knowledge‐driven Editing Paradigms}\label{sec-paradigm}

In \ourmethod, we propose three \emph{editing paradigms} to evaluate the effectiveness of different target knowledge for medical knowledge editing.

\noindent
\textbf{Ground‐Truth Answer Editing (GTA‐Edit)} is the prevailing paradigm that typically uses the final answer $a$ as the target knowledge $k$ for editing \citep{rome,memit}. As shown in \Cref{fig-overview}, GTA‐Edit directly injects the answer (e.g., ``Cervical vertebrae'') into LLMs. Although this approach is straightforward, it may lead to superficial memorization of the answer without a deep understanding of the underlying medical rationale.

\noindent
\textbf{Reference Editing (RE‐Edit)} takes supporting reference $c$ extracted from textbooks or academic literature as the target knowledge $k$ for editing. It provides a more comprehensive context to understand the underlying new knowledge.

\noindent
\textbf{Self‐generated Rationale Editing (SGR‐Edit)} 
is motivated by the rational postulates of belief revision \citep{logic}, 
which emphasizes that revision must be consistent and closed under logical consequence.
Moreover, prior work has argued that intelligent agents typically require an explanation to legitimize newly provided information, and the main role of an explanation is to rationalize facts \citep{logic2}.
Guided by these insights, we first prompt the LLM to generate a chain-of-thought rationale for the question-answer pair $(q, a)$ using the reference text $c$ as context. This generated rationale clearly outlines the reasoning process that leads to the answer, serving as the target knowledge $k$ for editing, as shown in the green part of \Cref{fig-overview}. Thus, SGR-Edit not only incorporates the answer but also provides a detailed explanation of how the model arrives at it, enabling LLMs to deliver interpretable and explainable reasoning that supports the new fact.
The detailed prompts for generating rationales are provided in Appendix \ref{app-sgr}. Examples of the GTA, RE, and SGR are presented in \Cref{diffk}.

\subsection{Evaluation Metrics}  
As shown in Figure~\ref{fig-overview}, the knowledge editing is conducted with questions $q\in\mathcal{Q}_{\mathrm{ori}}$ and their corresponding knowledge $k$ defined in each editing paradigm via \Cref{eq0} to inject knowledge into LLM.
The edited model $\theta'$ is then evaluated on three test sets, i.e., $q \in \mathcal{Q}_{\mathrm{ori}}\cup\mathcal{Q}_{\mathrm{gen}}\cup\mathcal{Q}_{\mathrm{ret}}$ to measure the \emph{efficacy}, \emph{generalization}, and \emph{retention} of each editing method with the accuracy of the predicted answer offered by the edited LLM.

To further evaluate the \emph{interpretability} of the edited model, we also instruct the edited LLM to provide a rationale for the final answer, which is compared with the ground-truth interpretation using ROUGE-L \citep{rouge}, BLEU scores \citep{bleu}, and Quality of Reasoning (QoR) scored via LLM-as-a-Judge. 
Detailed calculations are provided in \Cref{app:eval-metrics}.

\subsection{Editing Method Selection}

We choose six representative editing methods from three categories: \emph{Fine‐Tuning} methods - \textbf{LoRA} \citep{lora}, \emph{Parameter‐Modifying} methods - \textbf{ROME} \citep{rome} and \textbf{MEMIT} \citep{memit}, and \emph{Parameter‐Preserving} methods - \textbf{GRACE} \citep{grace}.  
Moreover, we include two newer \emph{Parameter‐Modifying} variants: \textbf{AnyEdit} \citep{anyedit} and \textbf{AlphaEdit} \citep{alphaedit}, which are state-of-the-art approaches supporting long-form and continual knowledge editing, respectively. 
Details of the KE methods are provided in Appendix \ref{app-methods}.

\subsection{Experimental Setup}

We evaluate KE methods on three LLMs: LLaMA-3.1-8B-Instruct, LLaMA-3.2-3B-Instruct\footnote{\url{https://huggingface.co/meta-llama}}, and Qwen2.5-7B-Instruct\footnote{\url{https://huggingface.co/Qwen}}. 
We utilize layers [4, 5, 6, 7, 8] for editing based on our findings from detailed experiments in Appendix \ref{sec-layers}.
Detailed experimental settings are provided in Appendix~\ref{app-set}.

\section{Main Experiments}

\begin{table*}[htbp]
  \centering
  \small
  \setlength{\tabcolsep}{2pt}
  \renewcommand{\arraystretch}{1.1} 
  \resizebox{0.75\linewidth}{!}{
  \begin{tabular}{ll c *{6}{c} } 
    \toprule
    \textbf{Method} & \textbf{Metric}
      & \textbf{Pre-Edit}
      & \multicolumn{2}{c}{\textbf{LLaMA-8B}}
      & \multicolumn{2}{c}{\textbf{LLaMA-3B}}
      & \multicolumn{2}{c}{\textbf{Qwen-7B}} \\ 
    \cmidrule(lr{.5em}){4-5} \cmidrule(lr{.5em}){6-7} \cmidrule(lr{.5em}){8-9} 
    & &
      & MedExQA\textsubscript{edit} & MedMCQA\textsubscript{edit}
      & MedExQA\textsubscript{edit} & MedMCQA\textsubscript{edit}
      & MedExQA\textsubscript{edit} & MedMCQA\textsubscript{edit} \\ 
    \midrule

    \multirow{4}{*}{LoRA}
      & Eff.  & 0   & 43.5 & 46.6 & 36.7 & 14.4 & 66.0 & 44.0 \\ 
      & Gen.  & 0   & 41.3 & 41.6 & 43.3 & 33.7 & 48.0 & 36.0 \\ 
      & Ret.  & 100 & 63.0 & 70.8 & 63.3 & 52.9 & 84.0 & 77.0 \\ 
      \cdashline{4-9} 
      & avg.  & \textbackslash
                    & \underline{49.3} & \underline{53.0} & \textbf{47.8} & 33.7 & \textbf{66.0} & \textbf{52.3} \\ 
    \specialrule{.75pt}{0pt}{0pt}

    \multirow{4}{*}{ROME}
      & Eff.  & 0   & 37.0 & 32.7 & 23.3 & 25.7 & 47.8 & 50.0 \\ 
      & Gen.  & 0   & 43.5 & 29.6 & 26.7 & 25.1 & 56.5 & 19.3 \\ 
      & Ret.  & 100 & 63.0 & 61.6 & 56.7 & 56.1 & 58.7 & 56.0 \\ 
      \cdashline{4-9} 
      & avg.  & \textbackslash
                    & 47.8 & 41.3 & 35.6 & 35.7 & 54.3 & 41.8 \\ 
    \specialrule{.75pt}{0pt}{0pt}

    \multirow{4}{*}{MEMIT}
      & Eff.  & 0   & 39.1 & 28.3 & 28.3 & 25.0 & 45.7 & 47.0 \\ 
      & Gen.  & 0   & 50.0 & 25.2 & 31.7 & 23.0 & 54.3 & 30.0 \\ 
      & Ret.  & 100 & 50.0 & 64.8 & 63.3 & 48.5 & 63.0 & 57.0 \\ 
      \cdashline{4-9} 
      & avg.  & \textbackslash
                    & 46.4 & 39.4 & 41.1 & 32.2 & 54.3 & 44.7 \\ 
    \specialrule{.75pt}{0pt}{0pt}

    \multirow{4}{*}{GRACE}
      & Eff.  & 0   & 34.8 & 36.0 & 33.3 & 26.7 & 39.1 & 23.0 \\ 
      & Gen.  & 0   & 21.7 & 29.2 & 10.0 & 7.0  & 43.5 & 32.0 \\ 
      & Ret.  & 100 & 76.1 & 80.7 & 88.3 & 93.0 & 63.0 & 69.0 \\ 
      \cdashline{4-9} 
      & avg.  & \textbackslash
                    & 44.2 & 48.7 & 43.9 & \underline{42.2} & 48.6 & 41.3 \\ 
    \specialrule{.75pt}{0pt}{0pt}

    \multirow{4}{*}{AnyEdit}
      & Eff.  & 0   & 34.8 & 36.6 & 35.0 & 23.0 & 39.1 & 26.0 \\ 
      & Gen.  & 0   & 28.3 & 25.5 & 28.3 & 23.0 & 52.2 & 27.0 \\ 
      & Ret.  & 100 & 78.3 & 78.9 & 71.7 & 78.1 & 73.9 & 83.0 \\ 
      \cdashline{4-9} 
      & avg.  & \textbackslash
                    & 47.1 & 47.0 & 45.0 & 41.4 & 55.1 & \underline{45.3} \\ 
    \specialrule{.75pt}{0pt}{0pt}

    \multirow{4}{*}{AlphaEdit}
      & Eff.  & 0   & 47.8 & 43.9 & 35.0 & 32.6 & 60.0 & 51.0 \\ 
      & Gen.  & 0   & 32.6 & 31.2 & 31.7 & 27.8 & 56.7 & 42.0 \\ 
      & Ret.  & 100 & 76.1 & 86.7 & 73.3 & 77.5 & 66.7 & 64.0 \\ 
      \cdashline{4-9} 
      & avg.  & \textbackslash
                    & \textbf{52.2} & \textbf{53.9} & \underline{46.7} & \textbf{46.0} & \underline{61.1} & \textbf{52.3} \\ 
    \specialrule{1.5pt}{0pt}{0pt}
  \end{tabular}%
  }
  \caption{Main results on medical knowledge editing with single editing (Accuracy \%).
    For avg. scores per column: \textbf{bold} is the best, \underline{underline} is the second best.}
  \label{tab:main_results}
\end{table*}

In this section, we evaluate the effectiveness of editing methods on medical knowledge by answering the following research questions:

\noindent
\textbf{RQ1:} How do current KE methods perform in the medical domain?

\noindent
\textbf{RQ2:} How do different KE paradigms impact the editing performance?

\noindent
\textbf{RQ3:} How does sequential medical KE affect the reliability and stability of in-domain knowledge?

\noindent
\textbf{RQ4:} Whether sequential medical KE affects broader domain knowledge?

\noindent
\textbf{RQ5:} How does sequential medical KE affect general LLM capabilities?

\subsection{Evaluating Editing Methods in Medical Domain (RQ1)}\label{sec-rq1}

Existing editing methods follow the ground‐truth answer editing (GTA‐Edit) paradigm, where the model is updated by injecting the answer $a$ as the target knowledge \citep{rome,memit}. We follow this paradigm to update LLMs with the ground‐truth answer $a$ for each question $q \in \mathcal{Q}_{\mathrm{ori}}$. Then, the model’s performance is evaluated on the three test sets $\mathcal{Q}_{\mathrm{ori}}\cup\mathcal{Q}_{\mathrm{gen}}\cup\mathcal{Q}_{\mathrm{ret}}$. From the results in \Cref{tab:main_results}, we observe that:

\paragraph{RQ1-F1: No existing editing method is effective enough for medical settings.}

The \emph{efficacy} of nearly all methods is below 50\%.
This sharply contrasts with previous reports of over 90\% efficacy in general domain benchmarks \citep{rome,alphaedit}, highlighting a significant gap when applying current KE methods to the intricate medical domain.

\paragraph{RQ1-F2: Existing editing methods struggle to generalize updated medical knowledge and often compromise existing knowledge.}  
This finding is reflected in the low \textit{generalization} scores and noticeable drops in \textit{retention}. For example, on MedMCQA\textsubscript{edit}, LoRA achieves only 41.6\% (8B) and 33.7\% (3B) generalization, while retaining 70.8\% and 52.9\% of previously correct answers.  
Despite its strong retention, GRACE shows limited generalization ability, making it less suitable for adapting to new clinical contexts.  
These results suggest that \textbf{existing editing paradigm (i.e., GTA-Edit) often leads to surface-level memorization rather than meaningful internalization of medical knowledge}.  
Detailed case analyses can be found in Appendix~\ref{app-cases}.

\subsection{Evaluation of Editing Paradigms (RQ2)}
\label{sec-rq2}

\begin{figure*}[!tb]
\centering
\includegraphics[width=0.78\textwidth]{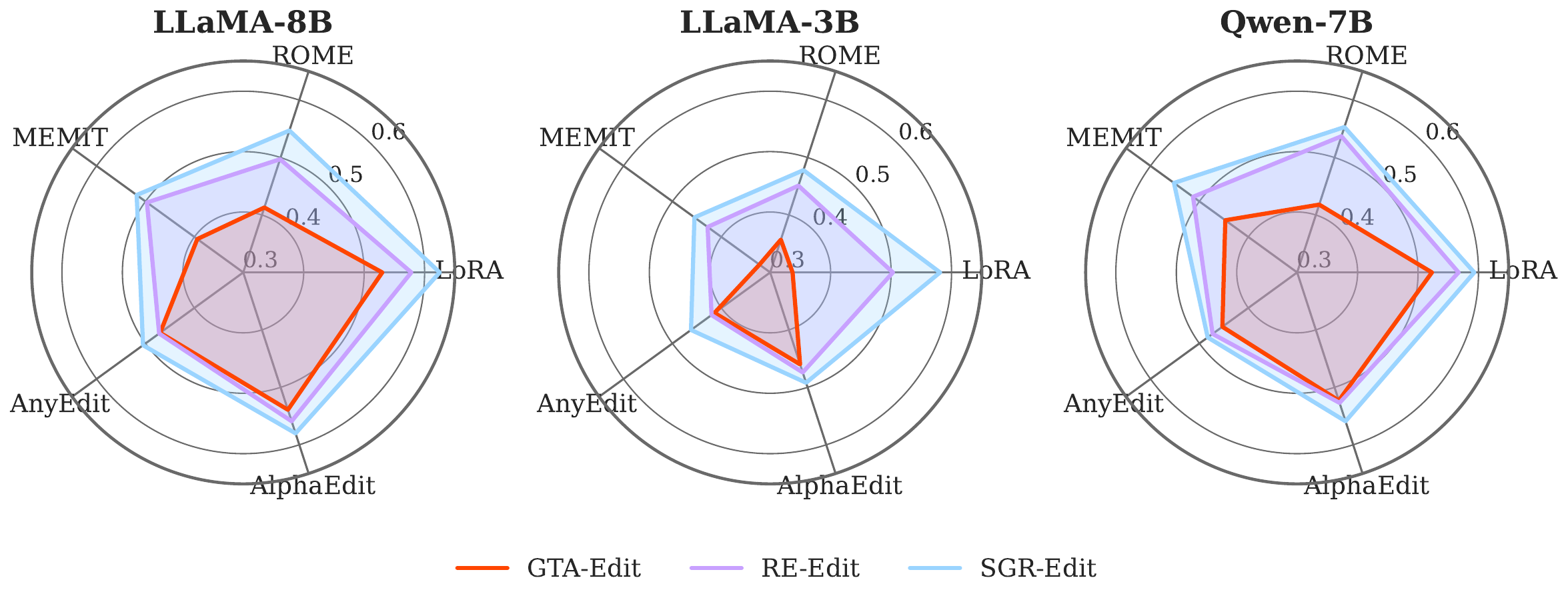}
\caption{Medical knowledge editing with various editing paradigms.}
\label{fig-radar}
\end{figure*}

This section investigates how different editing paradigms (i.e., GTA-Edit, RE-Edit, and SGR-Edit), affect the performance of existing editing methods.
To ensure the factual grounding of the rationales generated by the base LLM, we performed a targeted evaluation on a subset to assess rational factuality through a rigorous, two-stage process of AI-judge scoring and human verification (see Appendix \ref{sgr-check}).
Our key findings are as follows:

\paragraph{RQ2-F1: SGR-Edit yields the highest editing performance.}
As shown in Figure \ref{fig-radar}, SGR-Edit achieves the best performance for the five selected KE methods across various base LLMs.
Overall, SGR-Edit boosts performance by an average of 7.7\% relative to GTA-Edit on LLaMA-8B, by 11.3\% on LLaMA-3B, and by 6.3\% on Qwen-7B, demonstrating its consistently stronger capability across different model sizes and architectures.
Importantly, SGR-Edit achieves these gains with only one additional reasoning‐generation step, making it both highly effective and readily deployable. 
We discussed why SGR-Edit is practical in real-world employment in Appendix \ref{Practical-event}.
We also experimented with externally generated rationales in Appendix \ref{external-rationale}. 
The complete results and further analysis of various editing paradigms are provided in Appendix \ref{app-full}.  

\begin{figure*}[!htb]
\centering
\includegraphics[width=0.78\textwidth]{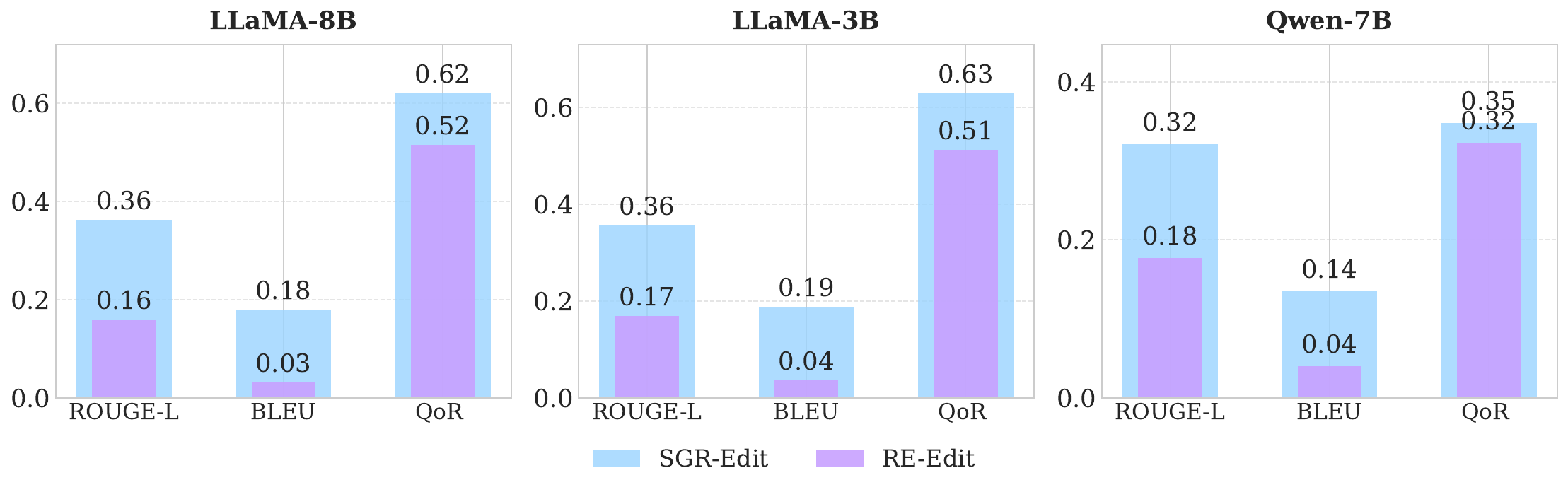}
\caption{Interpretability comparison for SGR-Edit and RE-Edit. Quality of Reasoning (QoR) is a normalized (0–1) score via an AI judge (DeepSeek-V3 \citep{deepseek}), assessing the generated rationales across five dimensions: factual accuracy, logical flow, relevance, completeness, and answer correctness (see prompt in \Cref{prompt:interpre}).}
\label{fig-interpre-memit}
\end{figure*}

\paragraph{RQ2-F2: SGR-Edit enables better reasoning interpretability.}
We compare the knowledge quality enabled by SGR-Edit and RE-Edit using MEMIT in Figure \ref{fig-interpre-memit}. SGR-Edit consistently and substantially outperforms RE-Edit across all models and metrics,
confirming that using the rationale-derived knowledge for editing results in deeper knowledge internalization. Full results for three KE methods are detailed in Appendix \ref{app-lex}.

\subsection{Sequential Editing Impacts on In-Domain Knowledge (RQ3)}
\label{sec-rq3}

\begin{figure*}[!htp]
\centering
\includegraphics[width=.85\textwidth]{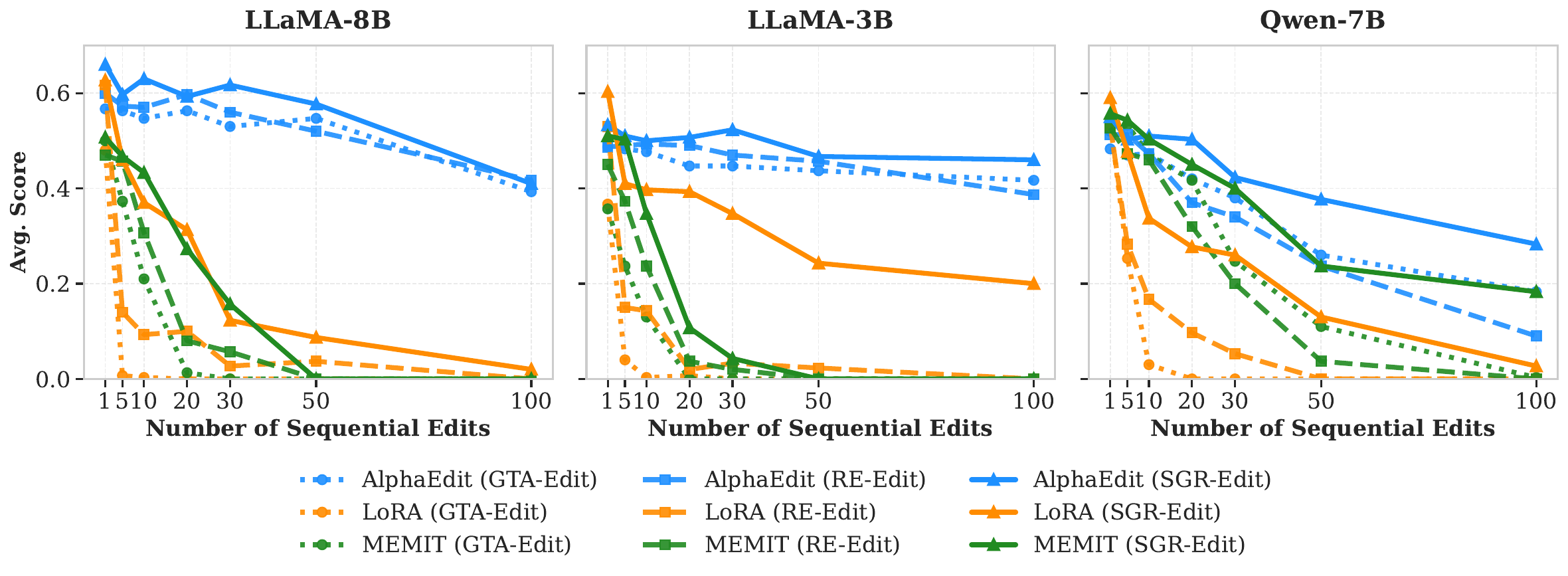}
\caption{Sequential editing performance on the internal benchmark.}
\label{fig-line}
\end{figure*}

We investigate how sequential editing affects the in-domain knowledge in LLMs, 
where knowledge items are continuously updated through multiple editing steps. 
Performance is tracked over \{1, 5, 10, 20, 30, 50, 100\} sequential edits using samples from \ourmethod, as shown in Figure~\ref{fig-line}.
We report our findings from three perspectives: \textbf{model-wise}, \textbf{method-wise}, and \textbf{paradigm-wise}:

\paragraph{RQ3-F1: Qwen experiences the most severe degradation under sequential edits, while LLaMA models exhibit greater stability.} 
All models experience gradual declines as the number of edits increases. 
However, the performance drop on Qwen-7B is substantially larger than that on LLaMA-8B or LLaMA-3B. 

\paragraph{RQ3-F2: Despite strong single-edit performance, LoRA suffers the sharpest decline under sequential edits, whereas AlphaEdit remains the most stable.} 
Across all models and paradigms (best viewed in color), LoRA quickly collapses after several edits, showing near-zero accuracy after 50–100 updates. 
MEMIT also degrades rapidly, while AlphaEdit sustains moderate performance even after 100 edits, highlighting its robustness in sequential editing.

\paragraph{RQ3-F3: Among editing paradigms, SGR-Edit is the most resistant to sequential degradation.} 
The rationale-driven SGR-Edit paradigm consistently preserves higher accuracy over repeated updates compared with GTA-Edit and RE-Edit (best viewed in line and marker). 
This indicates that SGR-Edit enables deeper knowledge integration and reduces catastrophic forgetting during continuous medical knowledge updates.

\subsection{Sequential Editing Impacts on External Domain Knowledge (RQ4)}\label{sec-rq4}

After assessing internal medical edits, we further investigate their influence on external-domain knowledge, covering both \textit{medical} and \textit{non-medical} aspects. 
We split the MMLU benchmark accordingly 
(see Appendix~\ref{app-mmlu-split} for subject mapping) 
to examine whether sequential medical editing propagates effects beyond the edited domain. Observations from Figure \ref{fig-seq-ext} are:

\begin{figure}[!htb]
\centering
\includegraphics[width=0.92\columnwidth]{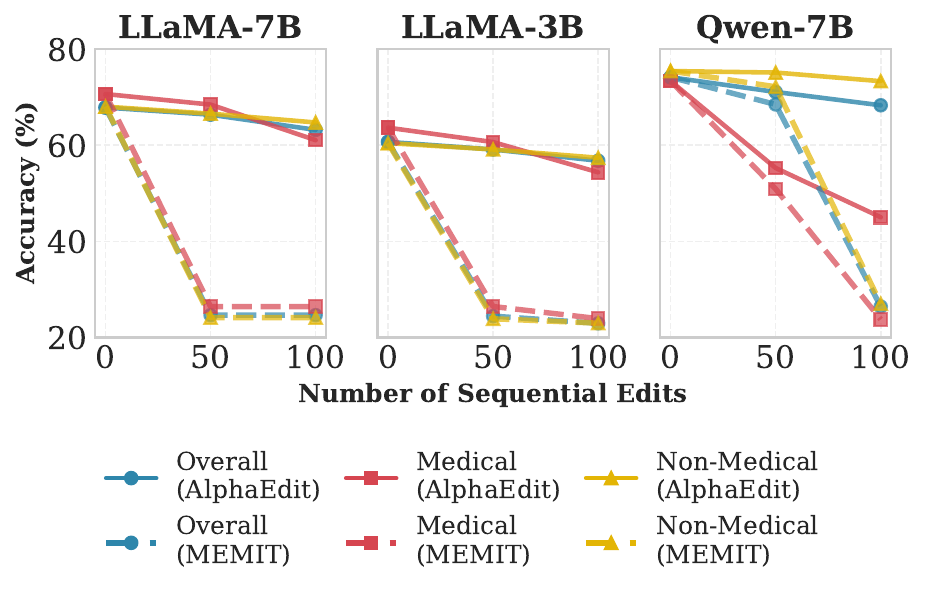}
\caption{Sequential editing performance on the external MMLU benchmark.}
\label{fig-seq-ext}
\end{figure}

\paragraph{RQ4-F1: Sequential medical edits consistently degrade both medical and non-medical knowledge from the external domain.}  
After 100 sequential edits using AlphaEdit on the internal \ourmethod, accuracy drops substantially not only in the medical domain from the MMLU benchmark — especially for Qwen-7B (-28.3\%) compared to LLaMA-8B (-9.5\%) and LLaMA-3B (-9.3\%), but also in non-medical categories (-3.3\% on LLaMA-8B and -3.1\% on LLaMA-3B).  
This indicates that sequential editing in internal medical knowledge erodes broader knowledge.


\paragraph{RQ4-F2: Advanced editing method mitigates external knowledge deterioration.} 
While both KE methods degrade performance, AlphaEdit consistently retains higher general-domain accuracies, as indicated by the solid lines, demonstrating that sequential optimization from the AlphaEdit mitigates catastrophic forgetting across domains.
We report category-wise results on MMLU general-domain tasks in Appendix Table~\ref{tab:rq4_common_domain_single}. 
We also provide detailed sequential editing results for traditional methods like ROME in Table~\ref{tab:rome_ext}, which exhibits similar catastrophic forgetting patterns to MEMIT, further confirming the necessity of sequential-optimized methods.
Further detailed comparison of different editing paradigms (GTA-Edit, RE-Edit, and SGR-Edit) and SGR edits with varying target knowledge lengths is presented in Table~\ref{tab:seq-sgr-len-tab}. 
We find that the primary driver of broader domain knowledge deterioration is the KE method (e.g., MEMIT's severe drop vs. AlphaEdit's minor drop), rather than the specific editing paradigm. Crucially, \textbf{increasing the target knowledge length in SGR-Edit does not cause significantly greater external knowledge degradation}.

\subsection{Sequential Editing Impacts on General LLM Capabilities (RQ5)}\label{sec-rq5}

\begin{table}[!htp]
\centering
\caption{Comparison of sequential editing performance on general benchmarks (AlphaEdit vs. MEMIT).}
\label{tab:combined_general_ability_rotated}
\resizebox{0.48\textwidth}{!}{%
\begin{tabular}{c l ccc ccc} 
\toprule
\multirow{2}{*}{\textbf{Model}} & \multirow{2}{*}{\textbf{Benchmark}} & \multicolumn{3}{c}{\textbf{\# Edits (AlphaEdit)}} & \multicolumn{3}{c}{\textbf{\# Edits (MEMIT)}} \\
\cmidrule(lr){3-5} \cmidrule(lr){6-8}
& & \textbf{0} & \textbf{50} & \textbf{100} & \textbf{0} & \textbf{50} & \textbf{100} \\
\midrule
\multirow{4}{*}{\rotatebox[origin=c]{90}{\textbf{LLaMA-8B}}} 
& HellaSwag & 79.2 & 78.8 & 78.5 & 79.2 & 26.2 & 25.9 \\
& TriviaQA & 51.9 & 50.1 & 48.4 & 51.9 & \textcolor{red}{0.0} & \textcolor{red}{0.0} \\
& TruthfulQA & 54.1 & 55.1 & 54.9 & 54.1 & 47.6 & 47.9 \\
& GSM8K & 77.9 & 76.7 & 76.5 & 77.9 & \textcolor{red}{0.0} & \textcolor{red}{0.0} \\
\midrule
\multirow{4}{*}{\rotatebox[origin=c]{90}{\textbf{LLaMA-3B}}} 
& HellaSwag & 70.4 & 70.7 & 70.0 & 70.4 & 26.2 & 26.5 \\
& TriviaQA & 33.8 & 31.6 & 27.6 & 33.8 & \textcolor{red}{0.0} & \textcolor{red}{0.0} \\
& TruthfulQA & 49.7 & 49.9 & 49.9 & 49.7 & 51.5 & 48.2 \\
& GSM8K & 64.6 & 65.4 & 62.2 & 64.6 & \textcolor{red}{0.0} & \textcolor{red}{0.0} \\
\midrule
\multirow{4}{*}{\rotatebox[origin=c]{90}{\textbf{Qwen-7B}}} 
& HellaSwag & 80.4 & 80.0 & 80.0 & 80.4 & 79.5 & 77.6 \\
& TriviaQA & 32.4 & 33.7 & 35.3 & 32.4 & 27.7 & 21.8 \\
& TruthfulQA & 64.7 & 63.6 & 63.2 & 64.7 & 63.4 & 56.5 \\
& GSM8K & 81.9 & 82.1 & 83.6 & 81.9 & 65.9 & 1.1 \\
\bottomrule
\end{tabular}
}
\end{table}

Building on our previous findings that SGR-Edit consistently outperforms its counterparts in both single and sequential medical editing, we further examine whether the edited LLMs can preserve their general capabilities. 
We evaluate them on four representative benchmarks: 
HellaSwag (commonsense reasoning), TriviaQA (factual recall), 
TruthfulQA (truthfulness) and GSM8K (mathematical reasoning).

\paragraph{RQ5-F1: Sequential-based KE method largely preserves general LLM capabilities, whereas the traditional KE method causes severe degradation and even model collapse.}
As shown in Table~\ref{tab:combined_general_ability_rotated}, the enhanced sequential editing AlphaEdit preserves most general capabilities even after 100 sequential edits,
remaining a more reliable solution for sustained medical knowledge updating while preserving the original LLM capabilities.
In contrast, MEMIT exhibits drastic degradation—its edited LLaMA models collapse completely on knowledge- and reasoning-intensive tasks 
(TriviaQA and GSM8K drop to 0.0). 
Upon inspection, these edited LLaMA models produce unreadable or nonsensical outputs (e.g., random punctuation), a failure pattern also reported in prior literature \citep{harm1,harm2}. 
This model collapse phenomenon is not unique to MEMIT. In Appendix~\ref{sec:seq-paradigms}, we present an extended analysis comparing AlphaEdit, ROME, and MEMIT across three editing paradigms (SGR, RE, GTA). The results confirm that traditional parameter-modifying methods (ROME, MEMIT) consistently suffer from catastrophic collapse in sequential settings, whereas AlphaEdit remains robust regardless of the editing paradigm.

\section{Conclusion}

This study introduces \ourmethod, a rigorous evaluation framework for medical knowledge editing designed to measure the reliability and generalization of updated knowledge in medical contexts. Our comprehensive evaluation reveals that the prevailing knowledge editing paradigm editing only simple final answers, often results in superficial memorization, failing to foster the deep internalization of medical principles necessary for generalization.
To address this limitation, we propose the novel KE paradigm called SGR-Edit, which targets the LLM's evidence-based reasoning chain instead of just the final answer. Extensive experiments demonstrate that SGR-Edit significantly boosts KE performance. Furthermore, our systematic analysis of sequential editing shows that the advanced KE method can effectively mitigate the degradation of broader knowledge and general abilities. These findings validate SGR-Edit as a robust KE strategy and provide crucial practical guidance for the sustainable deployment of explainable KE in real-world applications.

\section*{Limitations and Future Work}

We acknowledge the limitations of this study:

\begin{itemize}
    \item \textbf{Model Diversity}. We only evaluate two LLaMA-3 variants (3B, 8B) and Qwen2.5-7B in this study. Since our framework applies to any LLM by targeting its failure cases, broader validation on diverse and reasoning-enhanced models (e.g., Qwen3 \citep{qwen3}) is left for future work.
    
  \item \textbf{SGR-Edit Overhead:}  
    SGR-Edit requires only a single LLM and no external modules, leveraging evidence-based rationale generation without additional infrastructure. However, as the length of generated rationales grows, so does GPU memory consumption, which impedes large-batch editing experiments. Future work should investigate compact rationale representations to enable scalable batch updates.

  \item \textbf{Multi-Hop Reasoning:}  
    Real-world medical updates often involve interconnected facts and multi-step inferences. We leave the evaluation of multi-hop knowledge propagation and its downstream impact to future efforts.

  \item \textbf{Cross-Domain Generalization:}  
    While our primary focus is medical QA, the benchmark framework and SGR-Edit paradigm may generalize to other specialized domains (e.g., legal, scientific). We plan to assess and adapt our protocol for broader domain transferability.
\end{itemize}

\section*{Ethics Statement}

Our benchmarks are constructed from publicly available datasets and synthetic scenarios, which may introduce spurious or hallucinated content. 
Consequently, they are not for real clinical decision support.  
Furthermore, our findings highlight that editing operations can inadvertently degrade unrelated knowledge, underscoring the need for careful risk assessment before real-world deployment in safety-critical settings.  


\section*{Acknowledgments}
In this work, Shirui Pan was supported by an Australian Research Council (ARC) Future Fellowship (FT210100097) and DP240101547.
Carl Yang was not supported by any funds from China.

\bibliography{anthology,custom}

\appendix

\section{Medical Knowledge Editing Benchmarks}
\label{app-Benchmarks}

\subsection{Medical QA datasets}
\label{app-const}

\paragraph{MedMCQA}  
is drawn from postgraduate‐level Indian medical entrance exams (AIIMS and NEET PG), spanning 2,400 healthcare topics across 21 specialties. Each question offers four answer options \citep{medmcqa}. 
We construct a subset for MedMCQA\textsubscript{edit} by selecting up to 300 QA pairs from each specialty. After filtering out entries with missing explanations, multi-correct choices, or unknown subject labels, we obtain a high-quality set used to construct MedMCQA\textsubscript{edit}.

\paragraph{MedExQA}  
is designed to provide a richer medical context for evaluating LLMs by pairing each question with two human-curated explanation sets.
As these two explanations exhibit high semantic similarity (> 73\%) \citep{medexqa}, 
we simply uses the first explanation set in our benchmark.
It includes five underrepresented specialties in current datasets: biomedical engineering, clinical laboratory science, clinical psychology, occupational therapy, and speech language pathology.

We exclude the Speech Language Pathology subset on MedExQA due to frequent inconsistencies between questions and their provided explanations. 
For example, one multiple‐choice question lists B as the correct answer, while its accompanying explanation supports option D, rendering this subset unreliable for editing evaluations.
After removing Speech Language Pathology, the remaining four specialties comprise 773 QA pairs used to construct MedExQA\textsubscript{edit}.

\subsection{Data Construction}
\label{workflow}

Our benchmark construction proceeds in the follow four main steps to ensure that all evaluation questions truly measure editing gains in the medical domain:

\paragraph{Step 1. Quality Verification}  
We begin with two public medical QA datasets, MedExQA and MedMCQA, each of which pairs a question $q$ with a ground‐truth answer $k$ and one or two human‐written explanations $\mathrm{exp}$.  Due to known inconsistencies (e.g., mismatches between $k$ and $\mathrm{exp}$), we filter out any QA pair whose explanation fails to support the correct answer.  Concretely, we prompt an LLM (e.g., LLaMA-8B) with the $(\mathrm{exp},q)$ under in-context learning (more like in an open‐book setting).  If the model’s prediction does not match $k$, we discard that sample.  The remaining high‐quality pairs form our \emph{Verified Explanation Set} $\mathcal{D}_{\mathrm{verified}}$.

\paragraph{Step 2. Zero-Shot Filtering}  
Next, we assess this LLM’s zero‐shot performance on each question in $\mathcal{D}_{\mathrm{verified}}$ by feeding only $q$ (without $\mathrm{exp}$) in a context‐free condition.  Samples for which the model answers incorrectly indicate out-of-date or missing internal knowledge; these become our \emph{Original Set} $\mathcal{Q}_{\mathrm{ori}}$.

\begin{figure*}[!htb]
\centering
\includegraphics[width=1.0\textwidth]{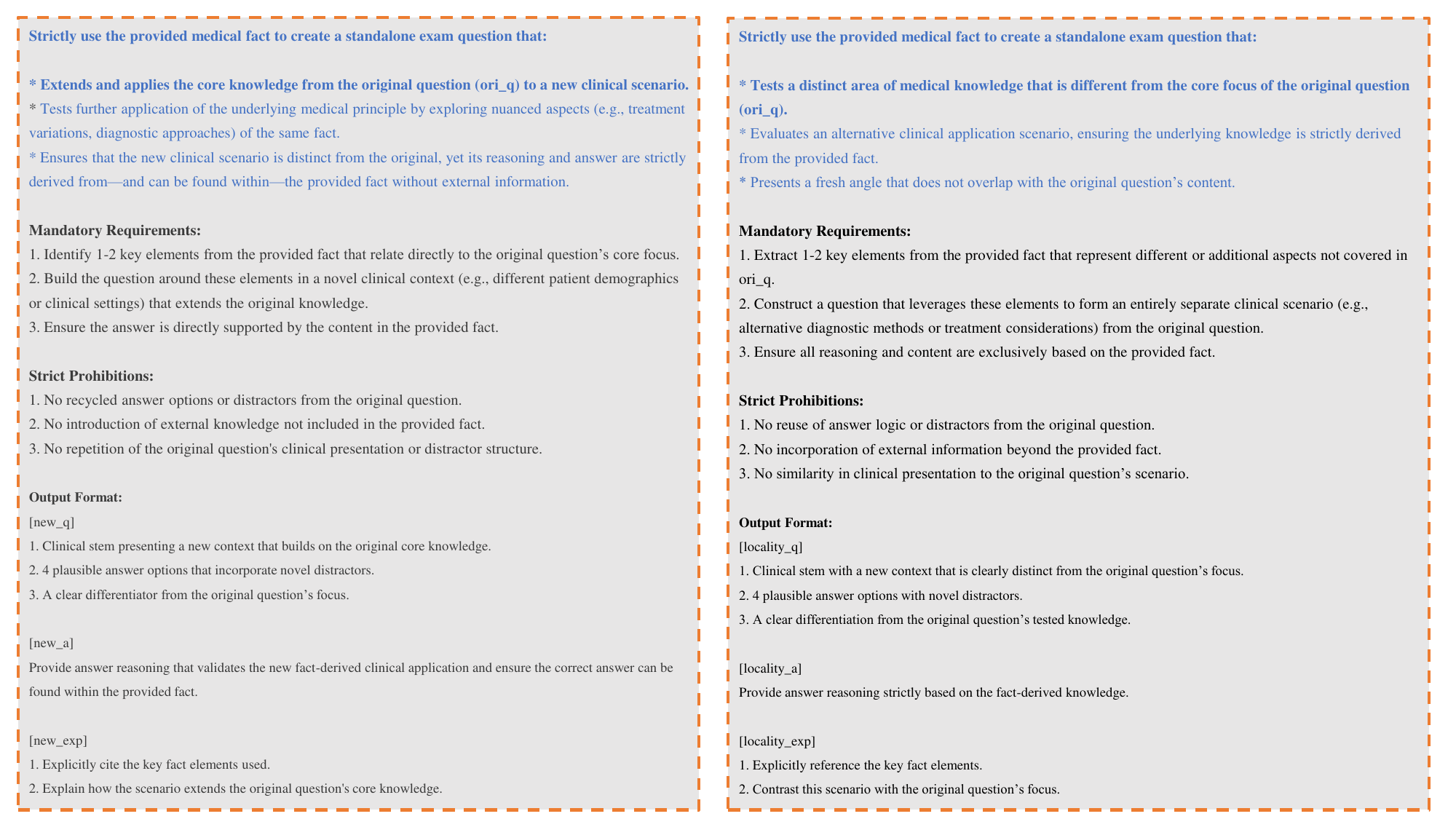}
  \caption{Prompts for $\mathcal{Q}_{\mathrm{gen}}$ (left) and $\mathcal{Q}_{\mathrm{ret}}$ (right) constructions}
\label{fig-newqa}
\end{figure*}

\paragraph{Step 3. Scenario Generation}  
To generate novel datasets for rigorous Generalization and Retention assessment, we use a more powerful agent (we use DeepSeek-V3 in this study; others like GPT-4 and Gemini could also serve this role) to construct new medical questions.
For each verified pair in $\mathcal{D}_{\mathrm{verified}}$, we treat its human‐curated explanation that cites authoritative medical textbooks, as the medical fact. 
Then, the agent uses the medical fact to craft clinical‐scenario QA pairs through prompting (see Figure \ref{fig-newqa}), yielding candidates for $\mathcal{Q}_{\mathrm{gen}}$ and $\mathcal{Q}_{\mathrm{ret}}$.  

\paragraph{Step 4. Data Filtering}
We then subject each generated variant to the same zero‐shot test as in Step 2.  Concretely, for generalization question candidates, we prompt the LLM without any explanation and record its predicted answer \(\hat{k}\).  If \(\hat{k} \neq k'\), where \(k'\) is the intended (ground‐truth) answer, we include it in the generalization set \(\mathcal{Q}_{\mathrm{gen}}\); 
As for retention question candidates, if \(\hat{k} = k'\), we add this pair to the retention set \(\mathcal{Q}_{\mathrm{ret}}\).  
This ensures that \(\mathcal{Q}_{\mathrm{gen}}\) contains only those scenarios that the LLM cannot solve without editing, while \(\mathcal{Q}_{\mathrm{ret}}\) captures instances where its original knowledge remains intact.

These processes yield three sets—$\mathcal{Q}_{\mathrm{ori}}$, $\mathcal{Q}_{\mathrm{gen}}$, and $\mathcal{Q}_{\mathrm{ret}}$, which together form our final editing benchmarks MedExQA\textsubscript{edit} and MedMCQA\textsubscript{edit}.

\begin{figure*}[!htb]
  \centering
  \includegraphics[width=\linewidth]{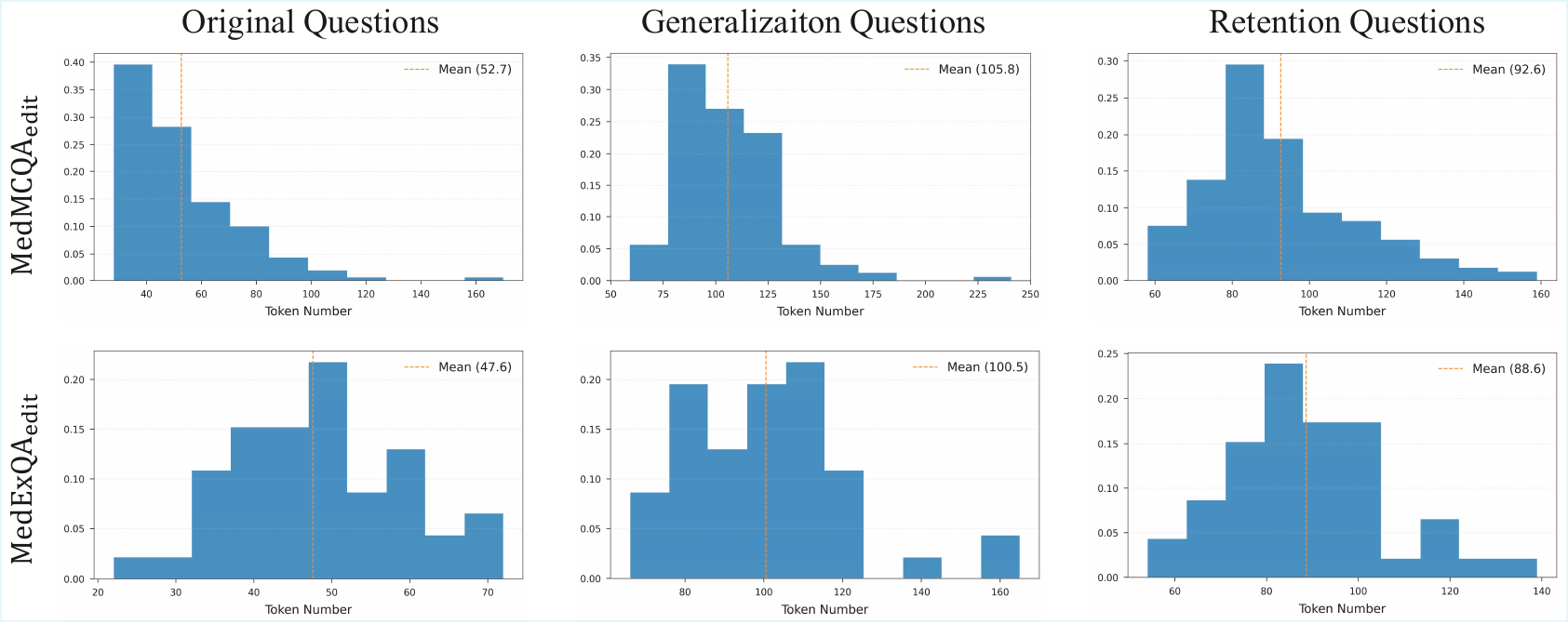}
  \caption{Token length distributions for MedMCQA\textsubscript{edit} (top row) and MedExQA\textsubscript{edit} (bottom row) across Original, Generalization, and Retention question sets. Dashed lines indicate dataset means. Our medical QA benchmarks present substantially longer inputs, increasing task difficulty.}
  \label{fig:token_dist}
\end{figure*}

\subsection{Benchmark Complexity}
\label{app-stat}

\begin{figure*}[ht]
  \centering
  \includegraphics[width=0.8\linewidth]{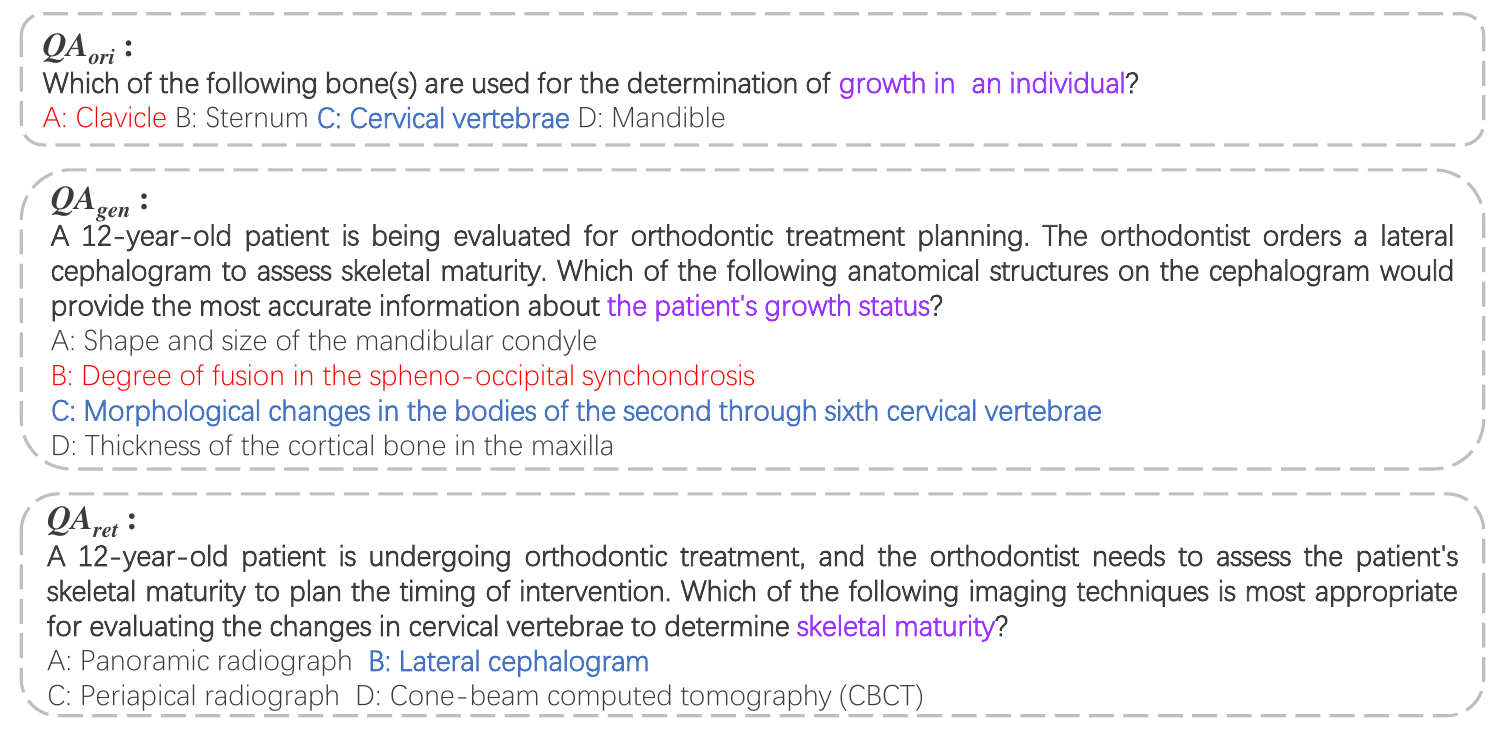}
  \caption{Examples of the three QA types. Purple text denotes the question’s domain/topic; blue text highlights the injected new correct answer; red text shows the model’s original (pre‑edit) answer.}
  \label{fig:QAs}
\end{figure*}

Our medical editing benchmarks introduce substantially more complex questions than common general‐domain datasets.  Whereas ZsRE questions average only 11.9 tokens, our MedMCQA\textsubscript{edit} and MedExQA\textsubscript{edit} samples exhibit mean lengths of 52.7 and 47.6 tokens for original questions, 105.8 and 100.5 for generalization questions, and 92.6 and 88.6 for retention questions, respectively (see Figure~\ref{fig:token_dist}).  This increase in question length reflects the inclusion of rich clinical context and novel scenario descriptions.  
In particular, $\mathcal{Q}_{\mathrm{gen}}$ and $\mathcal{Q}_{\mathrm{ret}}$ include extra patient‐centered details (see Figure \ref{fig:QAs}) to mimic real‐world clinical scenarios, further increasing the inference challenge. 
By expanding both the lexical and conceptual scope of each query, our benchmarks better simulate real‐world medical reasoning tasks and rigorously test an LLM’s capacity for knowledge editing under realistic complexity.

Steps 1 and 2 in Appendix \ref{workflow} are closely tied to how we interpret our evaluation metrics. If a base LLM were to answer all questions correctly or incorrectly, one of the evaluation sets would be empty, hindering a complete assessment of both efficacy and retention. In practice, when an LLM shows extreme pre-editing performance on a specific dataset or domain, we must dynamically adjust our focus: 

\begin{itemize}
    \item \textbf{If an LLM answers perfectly before editing, the evaluation would primarily focus on retention}, which measures how well the original knowledge is retained after editing, while efficacy becomes less critical as the target questions are already answered correctly.
    \item Conversely, \textbf{for a poorly performing LLM, efficacy becomes the dominant metric, focusing on the model's ability to correct its mistakes}.
\end{itemize}
This adaptive interpretation ensures a meaningful assessment across all performance spectra.

While our primary objective is to provide an efficient and automated framework for knowledge editing, we suggest that for scenarios demanding exceptionally high precision in sensitive real-world deployments, introducing “human-in-the-loop” verification for an additional layer of consistency checking would indeed be a beneficial and logical extension.

\section{Existing Editing Methods}
\label{app-methods}

\paragraph{Fine‐Tuning–Based Editing.}  
These traditional approaches update model parameters to incorporate new knowledge:
\textbf{FT+L} \citep{ftl} enforces norm constraints in gradient updates to minimize interference on the unmodified facts.
\textbf{FT‐M} \citep{ke1} applies a masking strategy during training to focus updates on relevant target content.
\textbf{LoRA} \citep{lora} introduces trainable low‐rank decomposition matrices to achieve efficient adaptation with minimal additional parameters.

\paragraph{Parameter‐Modifying Editing.}  
Unlike fine-tuning, these methods seek to modify only a subset of parameters to inject new facts while preserving unrelated knowledge. Broadly, they fall into two categories:
\textbf{i) Meta-Learning editing} typically trains a hypernetwork to estimate gradient updates for knowledge insertion \citep{ke,malmen}, e.g., \textbf{MEND} \citep{mend} introduces a hypernetwork to transform the gradient obtained by using a low-rank decomposition to make the parameterization tractable.
\textbf{ii) Locate‐then‐Edit} methods identify and then modify the weights responsible for specific factual associations:
      \textbf{ROME} \citep{rome} localizes and modifies factual associations in a transformer layer.  
      \textbf{MEMIT} \citep{memit} extends ROME by performing batch edits across multiple critical layers for mass knowledge updates.
      
Recently, sequential editing methods have been introduced to support continuous updates rather than a one-off modification.
    \textbf{PRUNE} \citep{prune} applies condition number constraints to limit perturbation to keep the general capacity of the model.  
    \textbf{AlphaEdit} \citep{alphaedit} minimizes disruption to the preserved knowledge by projecting perturbations onto the null space of its key matrices. 
    \textbf{AnyEdit} \citep{anyedit} decomposes long‐form knowledge into sequential chunks and edits each key token autoregressive. These three newer sequential editing methods are optimized on top of the MEMIT.

\paragraph{Parameter-Preserving Editing}  
These methods avoid altering the parameters of the base model.
    Extension‐based methods augment the model with external components, leaving the base parameters unchanged:
   \textbf{GRACE} \citep{grace} writes new mappings as entries in a discrete codebook in an Adaptor.
   \textbf{SERAC} \citep{serac} integrates a classifier and a side model to identify and learn new knowledge. Similarly, \textbf{WISE} \citep{wise} deploys side FFN layers to incorporate new knowledge dynamically.

  Additionally, retrieval‐based methods \citep{retrieval1,retrieval2,retrieval3} have been proposed to retrieve relevant factual information at inference time, effectively “editing” via context rather than weight changes. For example, \textbf{IKE} \citep{ike} retrieves relevant factual edits and uses them to build the prompt context as input, and then prompts the model to generate updated outputs.
  
\section{Detailed Experimental Settings}
\label{app-set}

\begin{table*}[!htb]
\small
\centering
\vspace{-6pt} 
\begin{tabularx}{\textwidth}{@{}>{\raggedright}p{1.2cm}X@{}}
\toprule
\textbf{Repr.} & \textbf{Content} \\
\midrule
GTA 
& “Co‐arctation of Aorta.” \\

\addlinespace[4pt]
RE
& \footnotesize
“Masson trichrome (Ref: Kanski 7/e p212–224; Parsons 22/e p212–214; Yanoff and Duker 4/e p261). Masson trichrome stain – Used for granular corneal dystrophy diagnosis. Granular dystrophy: AD inheritance with gene locus on 5q31; Onset: first decade with recurrent erosions; Signs: small, white, sharply demarcated deposits resembling crumbs or snowflakes in central anterior stroma; Histology: shows amorphous hyaline deposits staining bright red with Masson trichrome.” \\

\addlinespace[4pt]
SGR
& \footnotesize
“\textit{STEP 1:} \textcolor{blue}{According to the reference}, Masson trichrome stain is specifically used for the diagnosis of Granular dystrophy of the cornea.  
\textit{STEP 2:} The reference mentions that the histology of Granular dystrophy shows amorphous hyaline deposits staining bright red with Masson trichrome, \textcolor{blue}{indicating the effectiveness of this stain} in diagnosing the condition.  
\textit{STEP 3:} Colloidal iron stain is used for diagnosing various conditions, but…  
\textit{STEP 4:} Congo red stain is used for…  
\textit{STEP 5:} PAS (Periodic Acid–Schiff) stain is…” \\
\bottomrule
\end{tabularx}
\caption{Knowledge representations for editing. GTA: Bare correct answer; RE: Concise factual excerpt from expert source; SGR: Self-generated chain-of-thought rationale.}
\label{diffk}
\vspace{-5pt} 
\end{table*}

\subsection{Base Models}
In this study, we evaluate editing methods on two instruction‐tuned LLaMA variants: Llama-3.1-8B-Instruct and Llama-3.2-3B-Instruct \footnote{\url{https://llama.meta.com/lama3/}}. 
We utilize layers [4, 5, 6, 7, 8] for editing based on the findings from Appendix \ref{sec-layers}.

\subsection{Implementation}
We implement and evaluate six representative knowledge editing methods—LoRA, ROME, MEMIT, GRACE, AnyEdit, and AlphaEdit for fair comparison.
All editing workflows are built on the EasyEdit framework\footnote{\url{https://github.com/zjunlp/EasyEdit}}. For AnyEdit, which is not yet supported by EasyEdit, 
we integrate the original codebase\footnote{\url{https://github.com/jianghoucheng/AnyEdit}} and adopt the original hyperparameters.  
In terms of evaluation, we implement independent pipelines and consistently compute metrics tailored to our medical knowledge editing. 
During the post‐edit inference phase, we employ greedy decoding to ensure deterministic outputs. Specifically, we set \texttt{do\_sample=False} and \texttt{temperature=0.0} for all evaluations, so that the edited model’s predictions reflect its learned knowledge without sampling variability.  
For locate-and-edit KE methods, we select layers 4-8 as the target knowledge editing layers for all subsequent experiments (see Appendix \ref{sec-layers} for a detailed analysis).
For more specific experimental settings, please see our open-source codes.

\subsection{Editing Paradigm}
\label{app-sgr}

\begin{figure*}[!htb]
  \centering
  \includegraphics[width=\linewidth]{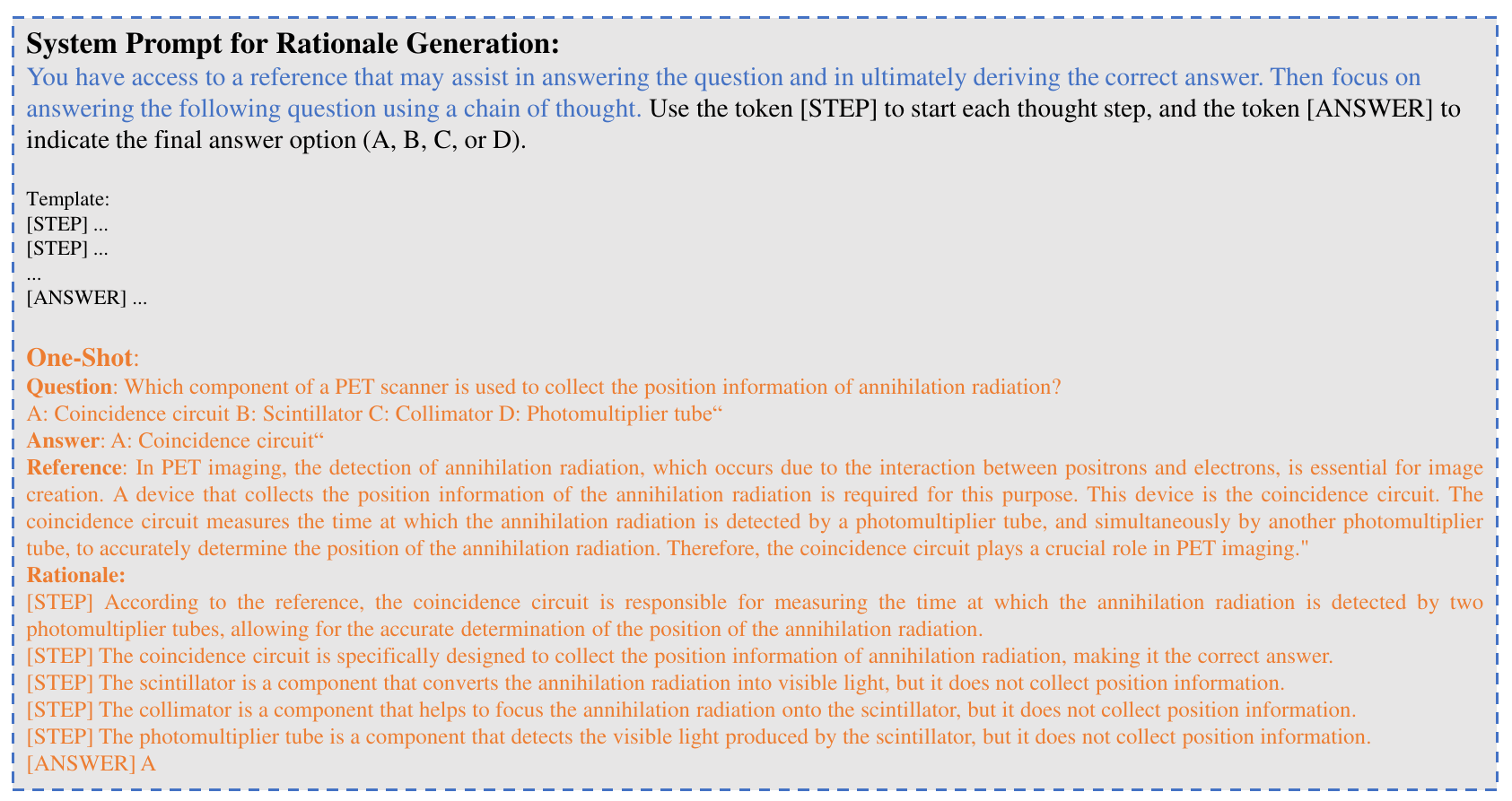}
  \caption{Complete prompt for generating rationales.}
  \label{fig:sgr-prompt}
\end{figure*}

For each $q_{\mathrm{ori}}\in\mathcal{Q}_{\mathrm{ori}}$, we construct three knowledge targets $k'$ according to the paradigm:
\begin{itemize}
  \item \textbf{GTA-Edit:} model is edited through a ground‐truth answer, with option letters stripped (e.g., remove “D:” in “D: Masson trichrome”).  
  \item \textbf{RE-Edit:} editing the model with the human‐written explanation excerpted from textbooks.  
  \item \textbf{SGR-Edit:} the proposed paradigm where the LLM is first prompted to generate its own chain of thought over the RE, then uses that self‐generated rationale as $k'$ for editing. The prompt used for rationale generation is in \Cref{fig:sgr-prompt}.
\end{itemize}

Examples of GTA, RE, and SGR are shown in Table \ref{diffk}.

\subsection{Evaluation Protocol}\label{app:eval-metrics}
During post‐edit evaluation, we randomly permute the answer choices in each multiple‐choice $\mathcal{Q}_{\mathrm{ori}}$ to prevent label memorization. All outputs follow a two‐step format: rationale generation followed by the final answer. We report:
\begin{itemize}
  \item \textbf{Efficacy:} accuracy on $\mathcal{Q}_{\mathrm{ori}}$.
  \item \textbf{Generalization:} accuracy on $\mathcal{Q}_{\mathrm{gen}}$.
  \item \textbf{Retention:} accuracy on $\mathcal{Q}_{\mathrm{ret}}$.
\end{itemize}

\noindent
\textbf{Notation.} For any set~$\mathcal{S}$, we write $\lvert \mathcal{S}\rvert$ for its cardinality (i.e., the number of elements in~$\mathcal{S}$).

\paragraph{Post‐edit Accuracy Metrics}
\begin{align}
\text{Efficacy} &= \frac{1}{\lvert \mathcal{Q}_{\mathrm{ori}}\rvert}
  \sum_{i \in \mathcal{Q}_{\mathrm{ori}}}
  \textbf{1}(\hat a_i = a_i), \\[4pt]
\text{Generalization} &= \frac{1}{\lvert \mathcal{Q}_{\mathrm{gen}}\rvert}
  \sum_{i \in \mathcal{Q}_{\mathrm{gen}}}
  \textbf{1}(\hat a_i = a_i), \\[4pt]
\text{Retention} &= \frac{1}{\lvert \mathcal{Q}_{\mathrm{ret}}\rvert}
  \sum_{i \in \mathcal{Q}_{\mathrm{ret}}}
  \textbf{1}(\hat a_i = a_i).
\end{align}

\noindent
\textbf{Interpretability:} ROUGE-L and BLEU scores between the injected knowledge (reference explanation or self-generated rationale) and the model’s post-edit rationale output, computed on a human-validated subset to quantify how closely the model reproduces the intended content.
Furthermore, we adopt Quality of Reasoning (QoR) as a normalized (0–1) score via an AI judge (DeepSeek-V3), assessing the generated rationales across five dimensions: factual accuracy, logical flow, relevance, completeness, and answer correctness (see prompt in \Cref{prompt:interpre})
\noindent Additional examples are provided in Appendix \ref{app-cases}.

\begin{figure*}[!htb]
  \centering
  \includegraphics[width=\linewidth]{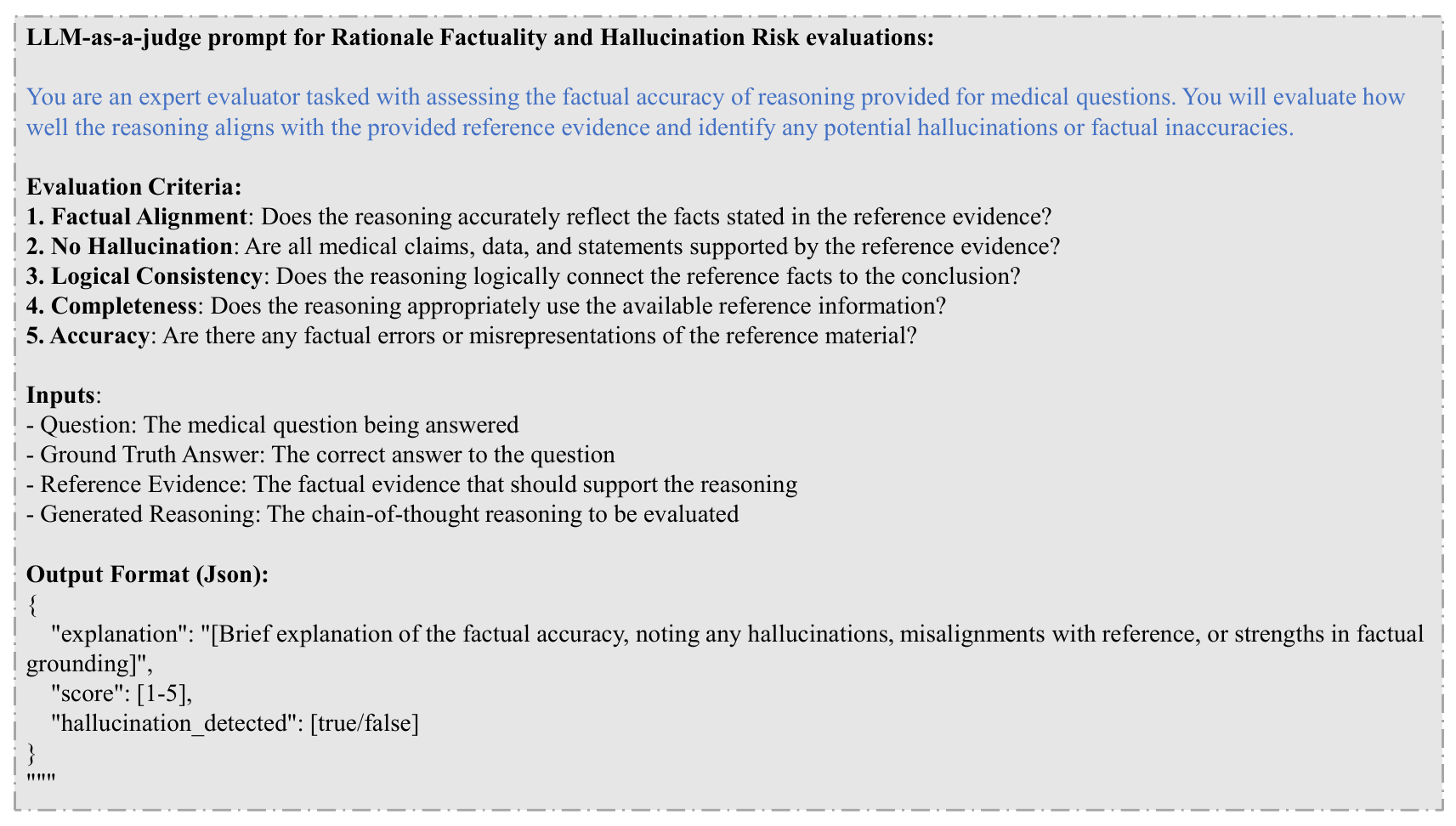}
  \caption{LLM-as-a-judge prompt for Rationale Factuality and Hallucination Risk evaluations}
  \label{hall}
\end{figure*}

\subsection{Task Formulation for Sequential Editing}

For $t$ sequential edits on distinct knowledge targets $\{(q_i, k_i)\}_{i=1}^t$, we define:
\begin{equation}\label{eq:sequential-edits}
\theta^{(i)} = F\bigl(\theta^{(i-1)},\,q_i,\,k_i\bigr), 
  i = 1,2,\dots,t
\end{equation}
to satisfy $\theta^{(i)}(q_i) = k_i,\ \forall\,i \in \{1,\dots,t\}$,
where $\theta^{(i)}$ is the model after the $i$-th edit.  After $t$ edits, the final model $\theta^{(t)}$ must satisfy $\theta^{(t)}(q_j) = k_j$ for all $ j\le t $
ensuring that each injected knowledge item remains correctly reflected in the model.

\section{Supplementary Results and Analyses}
\label{app-sup}

\subsection{Localization of Medical Knowledge in LLMs}
\label{sec-layers}

\begin{figure}[!htb]
\centering
\includegraphics[width=0.8\columnwidth]{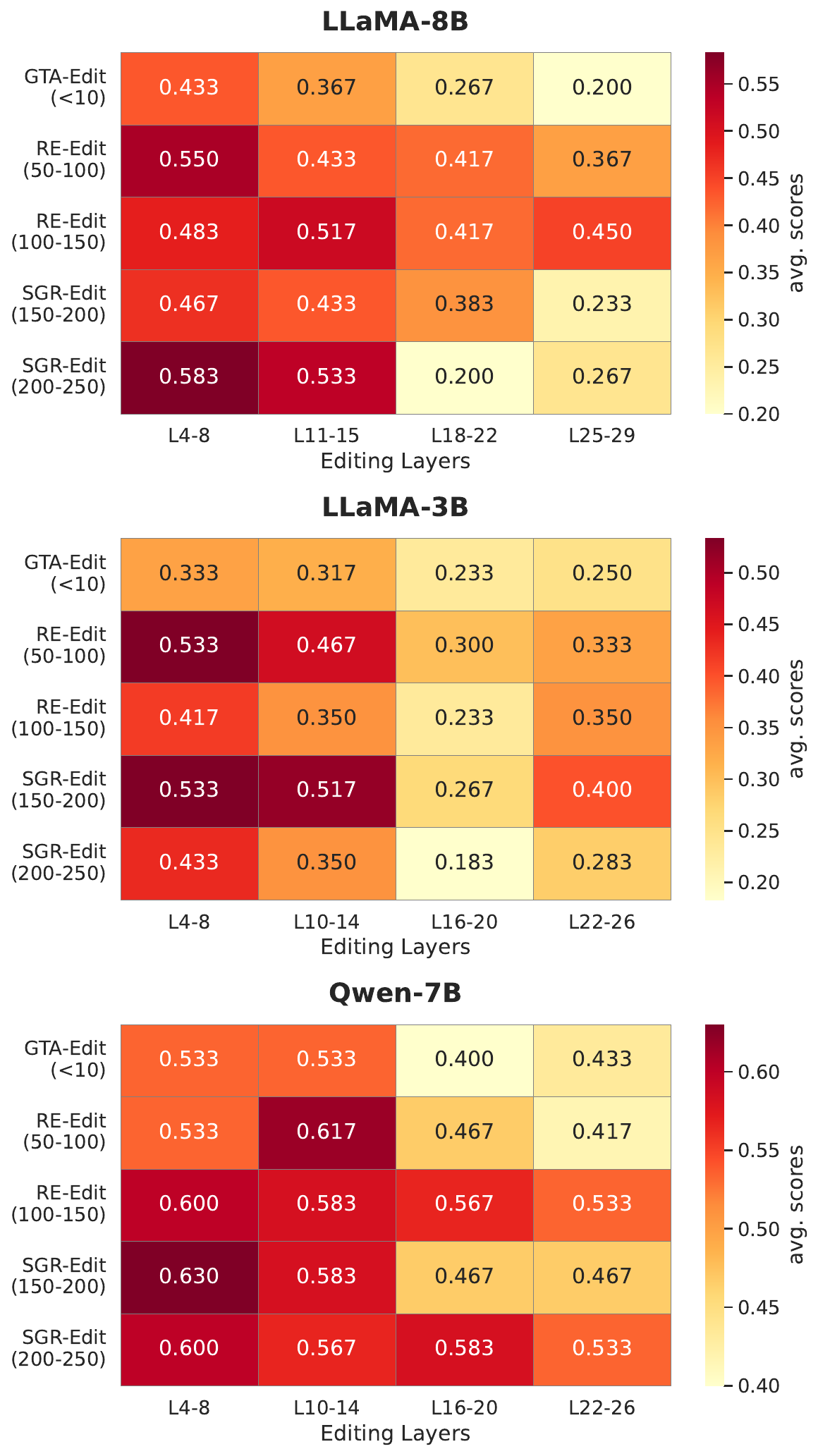}
\caption{Layer‐Wise editing performance}
\label{fig-hot}
\end{figure}
This section investigates the layer-wise storage of medical knowledge in LLMs. 
We edit knowledge across four disjoint layers to check the storage of certain knowledge: For LLaMA-8B, we select layers 4–8, 11–15, 18–22, and 25–29; for LLaMA-3B and Qwen-7B, we use layers 4–8, 10–14, 16–20, and 22–26.
We sample 100 QA pairs from MedMCQA\textsubscript{edit} and group target knowledge by token length to assess the impact of knowledge granularity. GTA targets are the shortest answers (<10 tokens), RE explanations span 50–100 or 100–150 tokens, and SGR rationales range from 150–200 to 200–250 tokens. Based on Figure~\ref{fig-hot}, we draw the following findings:

\paragraph{Medical knowledge is primarily stored in shallower layers.}
Unlike previous findings in general-domain settings, which suggest that factual associations are primarily stored in middle layers \citep{rome} and that deeper layers offer greater editing stability \citep{Navigating}, we find that medical knowledge in LLaMA models is most effectively edited in the shallower layers.

For both LLaMA-8B and 3B, editing operations targeting shallower layers 4-8 consistently yield the highest effectiveness. When editing on LLaMA-8B, average scores in layers 4-8 range from 0.433 to 0.583, and for LLaMA-3B, they span from 0.333 to 0.533, consistently outperforming edits in deeper layers.
Despite Qwen-7B also achieving strong editing effectiveness in deep layers (all above 40\%), the most effective network layers for it still reside in the shallower layers.

\paragraph{Across all editing paradigms, the most effective edits are consistently achieved in the shallower layers.}
For all target knowledge (i.e., GTA, RE, SGR), optimal editing performance is observed almost exclusively in layers 4-8. 
The results also indicate that while different target knowledge does not affect the optimal editing location, it does influence the magnitude of effectiveness.
For example, SGR-Edit (200-250) in layers 4-8 achieves the highest score of 0.583 for LLaMA-8B, with RE-Edit (50-100) demonstrating strong performance at 0.550 within these same layers. 

\noindent
Based on the above experiments and analysis, we adopt the shallower layers 4-8 as the knowledge editing layers for our experiments.

\subsection{Factual Grounding and Rationale Verification}
\label{sgr-check}

\begin{table*}[ht]
\centering
\resizebox{0.62\textwidth}{!}{
\begin{tabular}{lccccr}
\toprule
\textbf{Subset} & \textbf{Samples} & \textbf{Avg. Factual} & \textbf{Halluc.} & \textbf{Halluc.} \\
& & \textbf{Score} & \textbf{Count} & \textbf{Rate} \\
\midrule
MedExQA\textsubscript{edit} (LLaMA-3B) & 60 & 4.93 & 1 & 1.67\% \\
MedExQA\textsubscript{edit} (LLaMA-8B) & 46 & 5.00 & 0 & 0.00\% \\
MedExQA\textsubscript{edit} (Qwen-7B) & 50 & 4.90 & 2 & 4.00\% \\
\midrule
MedMCQA\textsubscript{edit} (LLaMA-3B) & 100 & 4.91 & 3 & 3.00\% \\
MedMCQA\textsubscript{edit} (LLaMA-8B) & 100 & 5.00 & 0 & 0.00\% \\
MedMCQA\textsubscript{edit} (Qwen-7B) & 100 & 4.99 & 0 & 0.00\% \\
\bottomrule
\end{tabular}
}
\caption{Factual evaluation and hallucination analysis across different subsets and models. The prompt is provided in Figure \ref{hall}}
\label{tab:factual_hallucination}
\end{table*}

To ensure the factual grounding of these rationales for both benchmark construction and the proposed editing paradigm, we performed a targeted evaluation on a subset to assess rational factuality. Our methodology involved two stages:

\begin{enumerate}
    \item AI Judge Evaluation: We first utilized DeepSeek-V3 as an AI judge to evaluate the generated rationales \citep{judge}. DeepSeek-V3 assessed the rationales across several dimensions, including Factual Alignment, Hallucination, Logical Consistency, Completeness, and Accuracy, ultimately assigning a factual score on a 1-5 scale and indicating whether hallucination was detected.
    \item Human Verification: Recognizing that even an advanced AI judge might exhibit its own form of hallucination in knowledge-specific tasks \citep{judge2}, especially with complex medical knowledge, we further introduced three human verifiers. These human annotators meticulously reviewed a subset of the rationales to confirm their factual accuracy and presence of hallucination, validating the AI judge's assessment.
\end{enumerate}

As observed from Table \ref{tab:factual_hallucination}, the average factual scores are remarkably high (ranging from 4.91 to 5.0), indicating strong factual alignment and consistency. Crucially, the hallucination rates are extremely low, with our rigorous verification process (involving both the AI judge and human annotators) confirming the high factual consistency of the generated rationales. This highlights that LLMs can self-generate high-quality rationales to update their knowledge.

\subsection{Event‐Driven Rationale Generation for Practical SGR‐Edit} 
\label{Practical-event}

In real‐world scenarios, knowledge updates are always triggered by concrete events: in medicine, for instance, the U.S. Food and Drug Administration’s approval of a novel oncology drug follows positive clinical trial results; in politics, a change in the presidency (e.g., from Biden to Trump) is driven by certified election outcomes; in law, the enactment of a new data‐privacy statute typically relies on high‐profile regulatory incidents. 
Under SGR-Edit, these domain-specific event narratives may be supplied by subject-matter experts or, alternatively, sourced automatically using Retrieval-Augmented Generation frameworks \citep{rag}.
By inputting these event narratives, the LLM can produce evidence‐grounded rationales that facilitate reliable and transparent knowledge edits in practice.

\subsection{External LLM for Rationale Generation}
\label{external-rationale}

We conducted an evaluation using a subset (100 samples per LLM), comparing performance when rationales are self-generated (SGR-Edit) by the base LLM versus when they are externally generated (EGR-Edit) by DeepSeek-V3. We applied these rationale types to ROME, MEMIT, and AlphaEdit methods, testing on LLaMA-8B and Qwen2.5-7B models.

\begin{table*}[ht]
\centering
\resizebox{0.52\textwidth}{!}{
\begin{tabular}{llcccc}
\toprule
\textbf{Method} & \textbf{Metric} & \multicolumn{2}{c}{\textbf{LLaMA-8B}} & \multicolumn{2}{c}{\textbf{Qwen2.5-7B}} \\
\cmidrule(lr){3-4} \cmidrule(lr){5-6}
& & \textbf{SGR-Edit} & \textbf{EGR-Edit} & \textbf{SGR-Edit} & \textbf{EGR-Edit} \\
\midrule
\multirow{4}{*}{\textbf{ROME}} & Eff. & 0.52 & 0.49 & 0.49 & 0.62 \\
& Gen. & 0.43 & 0.49 & 0.39 & 0.45 \\
& Ret. & 0.64 & 0.75 & 0.67 & 0.77 \\
\cline{2-6}
& \textbf{avg.} & 0.530 & \textbf{0.577} & 0.517 & \textbf{0.613} \\
\midrule
\multirow{4}{*}{\textbf{MEMIT}} & Eff. & 0.52 & 0.39 & 0.48 & 0.56 \\
& Gen. & 0.39 & 0.43 & 0.44 & 0.43 \\
& Ret. & 0.64 & 0.58 & 0.69 & 0.73 \\
\cline{2-6}
& \textbf{avg.} & \textbf{0.512} & 0.467 & 0.537 & \textbf{0.573} \\
\midrule
\multirow{4}{*}{\textbf{AlphaEdit}} & Eff. & 0.52 & 0.54 & 0.48 & 0.54 \\
& Gen. & 0.28 & 0.30 & 0.39 & 0.41 \\
& Ret. & 0.82 & 0.83 & 0.73 & 0.68 \\
\cline{2-6}
& \textbf{avg.} & 0.540 & \textbf{0.557} & 0.533 & \textbf{0.543} \\
\bottomrule
\end{tabular}
}
\caption{Editing performance comparison between SGR-Edit and ECG-Edit}
\label{tab:external-tab}
\end{table*}

Table \ref{tab:external-tab} shows that EGR-Edit generally yields higher average scores compared to SGR-Edit.
Despite this, our primary design philosophy for SGR-Edit was driven by crucial practical considerations,  especially for deployment in sensitive domains like medicine. \textit{SGR-Edit's core strength lies in enabling the local base LLM to self-update and refine its knowledge. This approach avoids the inherent security and privacy risks associated with distilling knowledge from or uploading sensitive local medical data to external LLMs/agents in real-world practice.} By relying on SGR-Edit, we maintain a self-contained and secure knowledge editing framework, which is paramount in contexts where data privacy and compliance are non-negotiable.

\begin{table*}[!htbp]
  \centering
  \small
  \setlength{\tabcolsep}{2.5pt} 
  \renewcommand{\arraystretch}{1.1}
  \resizebox{0.85\linewidth}{!}{
    \begin{tabular}{llc|ccc|ccc|ccc} 
      \toprule
      \multirow{2}{*}{\textbf{Method}} & \multirow{2}{*}{\textbf{Metric}} & \multirow{2}{*}{\textbf{Pre-Edit}}
          & \multicolumn{3}{c|}{\textbf{LLaMA-8B}}
          & \multicolumn{3}{c|}{\textbf{LLaMA-3B}}
          & \multicolumn{3}{c}{\textbf{Qwen-7B}} \\ 
      \cmidrule(lr){4-6} \cmidrule(lr){7-9} \cmidrule(lr){10-12} 
        & & & \textbf{GTA-Edit} & \textbf{RE-Edit} & \textbf{SGR-Edit}
          & \textbf{GTA-Edit} & \textbf{RE-Edit} & \textbf{SGR-Edit}
          & \textbf{GTA-Edit} & \textbf{RE-Edit} & \textbf{SGR-Edit} \\ 
      \midrule
      
      \multirow{4}{*}{LoRA}
        & Eff. & 0 & 0.466 & 0.547 & 0.665 & 0.144 & 0.412 & 0.615 & 0.440 & 0.540 & 0.590 \\ 
        & Gen. & 0 & 0.416 & 0.460 & 0.503 & 0.337 & 0.439 & 0.492 & 0.360 & 0.430 & 0.440 \\ 
        & Ret. & 1 & 0.708 & 0.727 & 0.708 & 0.529 & 0.658 & 0.636 & 0.770 & 0.730 & 0.750 \\ 
        & avg. & -- & 0.530 & 0.578 & \textbf{0.625} & 0.337 & 0.503 & \textbf{0.581} & 0.523 & 0.567 & \textbf{0.593} \\ 
      \midrule
      
      \multirow{4}{*}{ROME}
        & Eff. & 0 & 0.327 & 0.453 & 0.528 & 0.257 & 0.299 & 0.439 & 0.500 & 0.420 & 0.590 \\ 
        & Gen. & 0 & 0.296 & 0.403 & 0.447 & 0.251 & 0.406 & 0.385 & 0.193 & 0.510 & 0.320 \\ 
        & Ret. & 1 & 0.616 & 0.635 & 0.667 & 0.561 & 0.647 & 0.610 & 0.560 & 0.680 & 0.750 \\ 
        & avg. & -- & 0.413 & 0.497 & \textbf{0.547} & 0.357 & 0.451 & \textbf{0.478} & 0.418 & 0.537 & \textbf{0.553} \\ 
      \midrule
      
      \multirow{4}{*}{MEMIT}
        & Eff. & 0 & 0.283 & \textbf{0.435} & 0.520 & \textbf{0.250} & \textbf{0.345} & 0.455 & 0.470 & 0.480 & 0.575 \\ 
        & Gen. & 0 & 0.252 & \textbf{0.380} & 0.364 & \textbf{0.230} & \textbf{0.315} & 0.401 & 0.300 & 0.400 & 0.360 \\ 
        & Ret. & 1 & 0.648 & \textbf{0.675} & 0.671 & \textbf{0.485} & \textbf{0.625} & 0.663 & 0.570 & 0.660 & 0.720 \\ 
        & avg. & -- & 0.394 & 0.497 & \textbf{0.518} & 0.322 & 0.428 & \textbf{0.506} & 0.447 & 0.513 & \textbf{0.552} \\ 
      \midrule
      
      \multirow{4}{*}{GRACE}
        & Eff. & 0 & 0.366 & 0.410 & 0.435 & 0.230 & 0.246 & 0.316 & 0.230 & 0.230 & 0.230 \\ 
        & Gen. & 0 & 0.255 & 0.248 & 0.317 & 0.230 & 0.267 & 0.316 & 0.320 & 0.320 & 0.320 \\ 
        & Ret. & 1 & 0.789 & 0.758 & 0.764 & 0.781 & 0.747 & 0.754 & 0.690 & 0.690 & 0.690 \\ 
        & avg. & -- & 0.470 & 0.472 & \textbf{0.505} & 0.414 & 0.420 & \textbf{0.462} & 0.413 & 0.413 & \textbf{0.413} \\ 
      \midrule
      
      \multirow{4}{*}{AnyEdit}
        & Eff. & 0 & 0.366 & 0.410 & 0.435 & 0.230 & 0.246 & 0.316 & 0.260 & 0.380 & 0.370 \\ 
        & Gen. & 0 & 0.255 & 0.248 & 0.317 & 0.230 & 0.267 & 0.316 & 0.270 & 0.160 & 0.240 \\ 
        & Ret. & 1 & 0.789 & 0.758 & 0.764 & 0.781 & 0.747 & 0.754 & 0.830 & 0.880 & 0.840 \\ 
        & avg. & -- & 0.470 & 0.472 & \textbf{0.505} & 0.414 & 0.420 & \textbf{0.462} & 0.453 & 0.473 & \textbf{0.483} \\ 
      \midrule
      
      \multirow{4}{*}{AlphaEdit}
        & Eff. & 0 & 0.439 & 0.547 & 0.584 & 0.326 & 0.348 & 0.374 & 0.510 & 0.520 & 0.595 \\ 
        & Gen. & 0 & 0.312 & 0.335 & 0.366 & 0.278 & 0.332 & 0.348 & 0.420 & 0.360 & 0.362 \\ 
        & Ret. & 1 & 0.867 & 0.795 & 0.789 & 0.775 & 0.743 & 0.754 & 0.640 & 0.700 & 0.721 \\ 
        & avg. & -- & 0.539 & 0.559 & \textbf{0.580} & 0.460 & 0.474 & \textbf{0.492} & 0.523 & 0.527 & \textbf{0.559} \\ 
      \bottomrule
    \end{tabular}
    }
    \caption{Complete results of GTA-Edit, RE-Edit, and SGR-Edit across LLaMA-8B, LLaMA-3B, and Qwen-7B (Accuracy). SGR-Edit achieves the highest average score in nearly all comparisons.}
    \label{tab:para}
\end{table*}

\subsection{Performance Comparison of Various Editing Paradigms}
\label{app-full} 

As shown in Table \ref{tab:para}, relative to GTA-Edit, RE-Edit consistently raises average editing scores on LLaMA-8B by 0.2–8.4 percentage points across methods:
ROME sees the largest gain (+8.4 pp), followed by MEMIT (+5.3 pp), LoRA (+4.8 pp), AlphaEdit (+2.0 pp), and AnyEdit (+0.2 pp). These improvements are driven primarily by jumps in Efficacy (up to +12.6 pp for ROME) and Generalization (up to +10.7 pp for ROME), while Retention remains stable above 61\% for all methods.
Incorporating SGR-Edit yields further average gains of 2.1–7.1 pp over RE-Edit, with MEMIT (+7.1 pp) and LoRA (+4.7 pp) benefiting most. Specifically, LoRA’s combined score climbs from 57.8\% to 62.5\%, and MEMIT from 44.7\% to 51.8\%. Even AlphaEdit, which already excels under RE-Edit, improves from 55.9\% to 58.0\% (+2.1 pp). 
Critically, these gains in Efficacy and Generalization come with only minor retention trade-offs (e.g., AnyEdit drops from 78.9\% to 76.4\%), confirming that richer and context-driven rationales enable deeper medical knowledge integration without undue forgetting. 

The observed trend of improvement from RE-Edit to SGR-Edit extends robustly to the Qwen-7B model. Across all methods on Qwen-7B, SGR-Edit achieves the highest average editing scores, demonstrating its superiority regardless of the base LLM architecture.
Notably, ROME and MEMIT exhibit substantial gains, increasing their average scores by +13.5 pp and +10.5 pp from GTA-Edit to SGR-Edit, respectively, confirming that the enriched SGR target benefits knowledge integration in Qwen-7B as well.

Additionally, we note that the GRACE method exhibited an anomalous pattern across multiple trials: its performance remained nearly identical across all three editing paradigms (i.e., GTA-Edit, RE-Edit, and SGR-Edit). We speculate this is due to GRACE's reliance on a codebook's deferral radius, limiting edits to inputs near existing keys, which makes it less sensitive to the content variation. We leave this phenomenon for future investigation.

\begin{figure*}[!htb]
\centering
\includegraphics[width=0.74\textwidth]{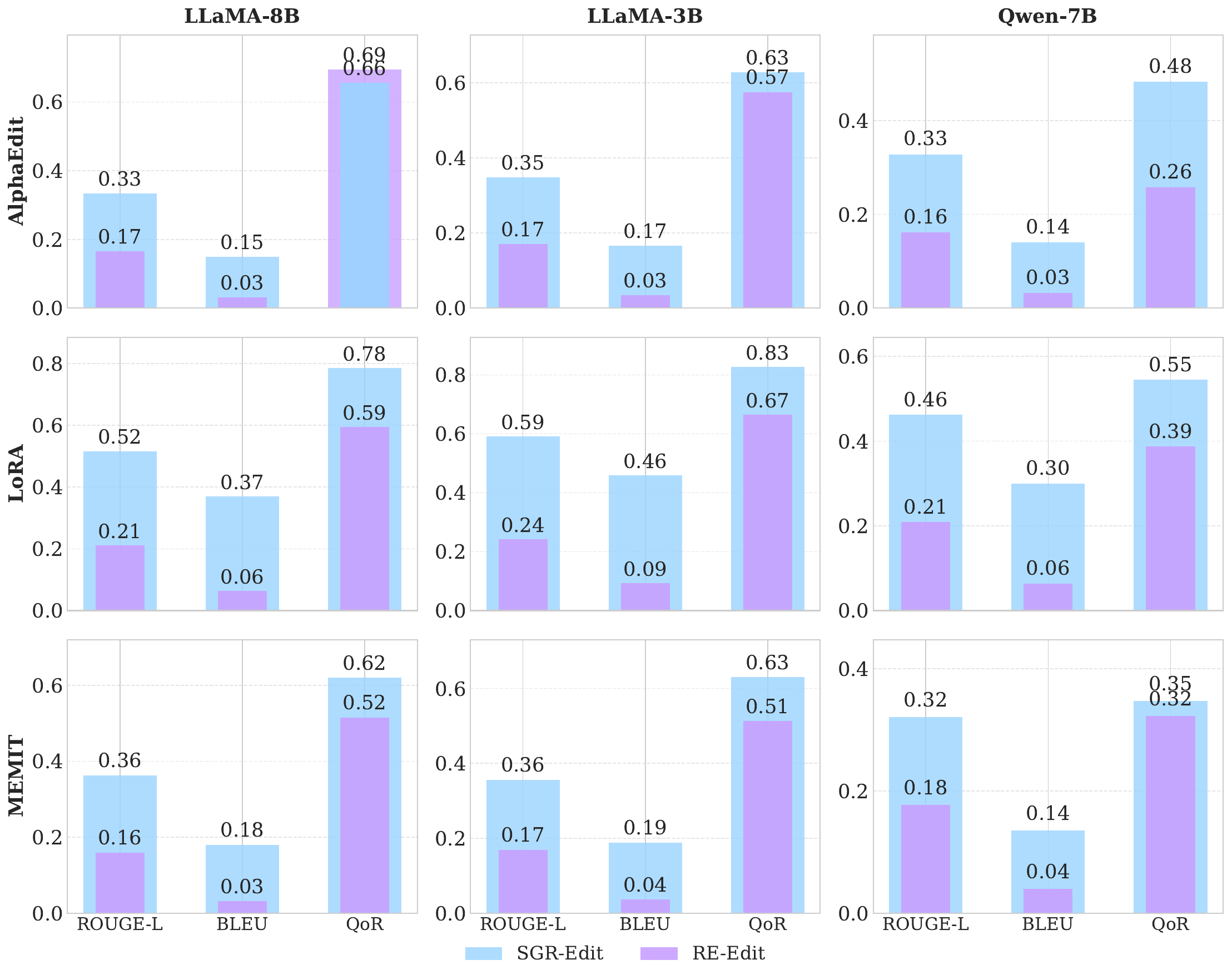}
\caption{Interpretability comparison for SGR-Edit and RE-Edit using AlphaEdit, LoRA, and MEMIT.
The metric \textbf{QoR} (Quality of Reasoning) represents a normalized score (0 to 1) obtained from an LLM-as-a-Judge (DeepSeek-V3) evaluation, assessing answer accuracy, medical factual accuracy, logical flow, relevance, and completeness.}
\label{fig-bar}
\end{figure*}

\subsection{Interpretability Analysis of SGR-Edit and RE-Edit}
\label{app-lex} 

The results in Figure~\ref{fig-bar} are computed over a human‐validated MedMCQA\textsubscript{edit} subset of high‐quality reference explanations (RE) and self‐generated rationales (SGR). Since SGR generation can include spurious content, we manually verify each rationale (see Table \ref{tab:factual_hallucination}) to ensure it faithfully supports the target answer before using it for ROUGE-L and BLEU calculations.
These long‐form explanations provide rich contextual support for question answering, enabling a precise evaluation of lexical overlap between the injected knowledge and the model’s post‐edit outputs. 

Across all three editing methods: AlphaEdit, LoRA, and MEMIT, SGR-Edit outperforms RE‐Edit by a substantial margin in both ROUGE‐L and BLEU. For LLaMA‐8B, the average lexical overlap (see Lexical avg.) for SGR‐Edit is 0.241 (AlphaEdit), 0.443 (LoRA), and 0.271 (MEMIT), compared to just 0.098, 0.137, and 0.095 under RE‐Edit. 
BLEU improvements are equally dramatic: AlphaEdit rises from 0.031 to 0.149, LoRA from 0.064 to 0.370, and MEMIT from 0.031 to 0.180. These gains confirm that SGR—Edit allows evidence‐grounded and logical knowledge representation to align more closely with the correct rationale for medical decision making and thus serve as a superior knowledge target.

When comparing across editing methods, LoRA consistently achieves the highest lexical overlap among all paradigms and model sizes. On LLaMA‐8B, LoRA SGR‐Edit reaches ROUGE‐L=0.516 and BLEU=0.370, yielding a text‐average of 0.443. This outperforms both AlphaEdit (0.241 lexical avg.) and MEMIT (0.271 lexical avg.), indicating that LoRA’s low‐rank adaptation effectively internalizes the rich and context‐driven rationales generated by the model. The pattern also holds on LLaMA‐3B, where LoRA SGR‐Edit achieves a lexical average score of 0.526 versus 0.257 (AlphaEdit) and 0.272 (MEMIT).

\begin{figure*}[!htb]
\centering
\includegraphics[width=0.9\textwidth]{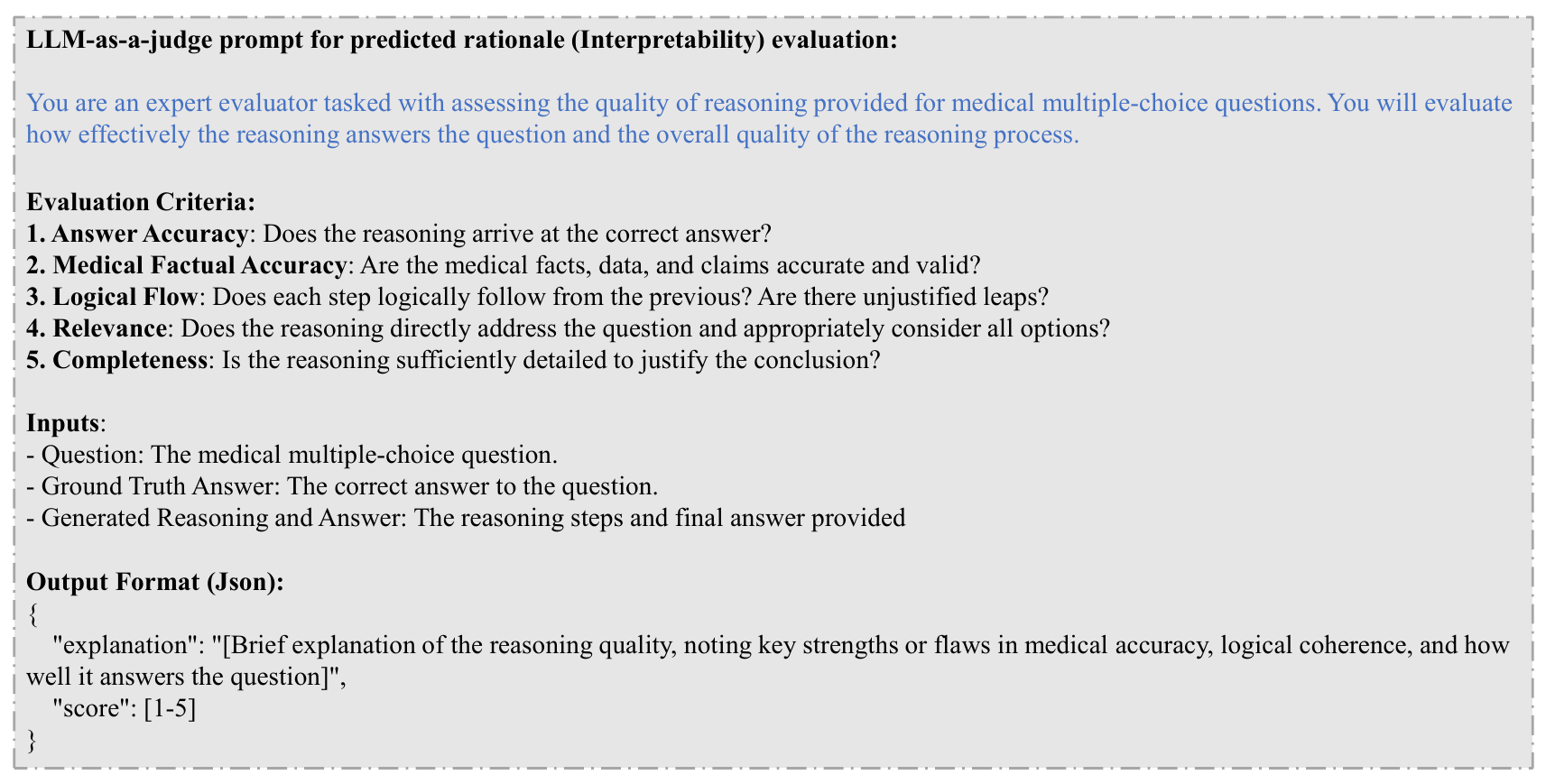}
\caption{LLM-as-a-judge prompt for predicted rationale (Interpretability) evaluation}
\label{prompt:interpre}
\end{figure*}

We further analyze the overall quality of the generated rationales using the Quality of Reasoning (QoR) Score. This is achieved by prompting the DeepSeek-V3 to evaluate output quality across several dimensions: Answer Accuracy, Medical Factual Accuracy, Logical Flow, Relevance, and Completeness (the prompt are provided in Figure \ref{prompt:interpre}). SGR-Edit generally yields a higher Quality Score than RE-Edit, suggesting that using the model's self-generated context results in more cohesive and contextually rich explanations.

In summary, these metrics corroborate our finding that SGR‐Edit consistently yields the highest interpretability, validating its ability to convey a deeper understanding rather than superficial information memorization. Furthermore, LoRA emerges as the most effective editing mechanism, capitalizing on the enriched content of self‐generated rationales to maximize knowledge integration.

\subsection{Post-Edit Inference: Two-Step Rationale \& Answer vs. One-Step Final Answer}
\label{app-output}

\begin{figure}[!htb]
\centering
\includegraphics[width=0.7\columnwidth]{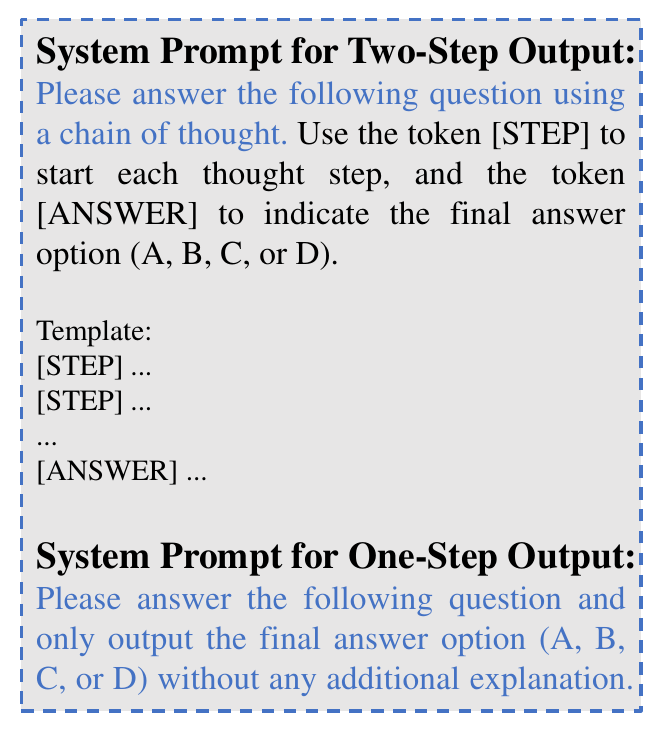}
\caption{Two-Step and One-Step System Prompts for Post-Edit Inference}
\label{fig-post}
\end{figure}

\begin{table*}[ht]
    \centering
    \small
    \resizebox{0.82\textwidth}{!}{
    \begin{tabular}{llc|ccc|ccc}
        \toprule
        \multirow{2}{*}{\textbf{Method}} & \multirow{2}{*}{\textbf{Metric}} & \multirow{2}{*}{\textbf{Pre-Edit}}
            & \multicolumn{3}{c|}{\textbf{Final-Answer Only}}
            & \multicolumn{3}{c}{\textbf{Rationale + Answer}} \\
        \cmidrule(lr){4-6} \cmidrule(lr){7-9}
         &  &  & \textbf{GTA-Edit} & \textbf{RE-Edit} & \textbf{SGR-Edit}
               & \textbf{GTA-Edit} & \textbf{RE-Edit} & \textbf{SGR-Edit} \\
        \midrule
        \multirow{4}{*}{LoRA}
         & Eff. & 0  & 0.296 & 0.553 & 0.616  & 0.466 & 0.547 & 0.665 \\
         & Gen. & 0  & 0.220 & 0.258 & 0.258  & 0.416 & 0.460 & 0.503 \\
         & Ret. & 1  & 0.465 & 0.591 & 0.453  & 0.708 & 0.727 & 0.708 \\
         & avg. & -- & 0.327 & 0.468 & 0.442  
                    & 0.530\ \textcolor{blue}{(+0.203)}
                    & 0.578\ \textcolor{blue}{(+0.110)}
                    & 0.625\ \textcolor{blue}{(+0.183)} \\
        \midrule
        \multirow{4}{*}{ROME}
         & Eff. & 0  & 0.528 & 0.566 & 0.642  & 0.327 & 0.453 & 0.528 \\
         & Gen. & 0  & 0.252 & 0.277 & 0.277  & 0.296 & 0.403 & 0.447 \\
         & Ret. & 1  & 0.484 & 0.491 & 0.472  & 0.616 & 0.635 & 0.667 \\
         & avg. & -- & 0.421 & 0.444 & 0.463  
                    & 0.413\ \textcolor{red}{(-0.008)}
                    & 0.497\ \textcolor{blue}{(+0.053)}
                    & 0.547\ \textcolor{blue}{(+0.084)} \\
        \midrule
        \multirow{4}{*}{MEMIT}
         & Eff. & 0  & 0.440 & 0.541 & 0.566  & 0.283 & 0.384 & 0.520 \\
         & Gen. & 0  & 0.214 & 0.270 & 0.302  & 0.252 & 0.352 & 0.364 \\
         & Ret. & 1  & 0.509 & 0.484 & 0.484  & 0.648 & 0.604 & 0.671 \\
         & avg. & -- & 0.388 & 0.432 & 0.451  
                    & 0.394\ \textcolor{blue}{(+0.006)}
                    & 0.447\ \textcolor{blue}{(+0.015)}
                    & 0.518\ \textcolor{blue}{(+0.067)} \\
        \midrule
        \multirow{4}{*}{AlphaEdit}
         & Eff. & 0  & 0.623 & 0.654 & 0.648  & 0.439 & 0.547 & 0.584 \\
         & Gen. & 0  & 0.252 & 0.270 & 0.233  & 0.312 & 0.335 & 0.366 \\
         & Ret. & 1  & 0.648 & 0.623 & 0.635  & 0.867 & 0.795 & 0.789 \\
         & avg. & -- & 0.507 & 0.516 & 0.505  
                    & 0.539\ \textcolor{blue}{(+0.032)}
                    & 0.559\ \textcolor{blue}{(+0.043)}
                    & 0.580\ \textcolor{blue}{(+0.075)} \\
        \bottomrule
    \end{tabular}}
    \caption{Post-Edit accuracy comparison between Final‐Answer Only and Rationale + Answer}
    \label{tab:postedit_accuracy_comparison}
\end{table*}

We compare two prompting strategies on LLaMA-8B over MedMCQA\textsubscript{edit}: (i) “Rationale + Answer”  and (ii) “Final‐Answer Only” (see Figure~\ref{fig-post}). Table~\ref{tab:postedit_accuracy_comparison} presents post‐edit accuracy under GTA-Edit, RE-Edit, and SGR-Edit for LoRA, ROME, MEMIT, and AlphaEdit. Overall, providing a chain of thought during post-edit inference yields substantial gains in editing efficacy and generalization, at minimal retention cost.

For LoRA, the average post‐edit score climbs from 0.442 to 0.530 (+0.203) when moving to two‐step output. Efficacy improves dramatically (GTA-Edit: +0.170, RE-Edit: +0.106, SGR-Edit: +0.049), and Generalization more than twice in the GTA-Edit case (0.220 $\rightarrow$ 0.416). Retention also rises under RE (0.591 $\rightarrow$ 0.727), demonstrating that transparent reasoning reinforces new facts.

ROME shows a mixed pattern: while GTA-Edit average performance drops slightly (0.421 $\rightarrow$ 0.413), RE-Edit and SGR-Edit see gains (+0.053 and +0.084 avg.), reflecting that rationale prompts help when richer contexts are available. 
MEMIT benefits modestly (+0.006 to +0.067 avg.), with Generalization under SGR especially boosted (0.302 $\rightarrow$ 0.364). 
AlphaEdit gains across all paradigms (+0.032 to +0.075 avg.), with Retention remaining above 0.789 even after two‐step reasoning.

These results confirm that explicit chain‐of‐thought prompting significantly enhances the model’s ability to apply the injected knowledge, supporting more reliable and interpretable medical knowledge editing.

\subsection{Sequential Editing Impact on Common‐Domain Capabilities}
\label{app-mmlu-split}

We split the MMLU benchmark into medical and non‐medical categories: the medical sets are computed over the subjects \{anatomy, clinical\_knowledge, college\_medicine, human\_aging, medical\_genetics, nutrition, professional\_medicine, virology\}, which are typically referred to as \emph{health-related} in the MMLU community.
For consistency with our medical evaluation setting, we collectively denote them as the \textbf{medical} subset, while all remaining subjects are aggregated as the \textbf{non-medical} subset.

In addition to the medical vs. non‐medical breakdown (see section~\ref{sec-rq4}), we further split MMLU into four broad categories—STEM, Humanities, Social Sciences, and Other (Business, Misc.)—to assess general‐domain degradation. Table~\ref{tab:rq4_common_domain_single} reports category accuracies for LLaMA‐8B and LLaMA‐3B at 0, 50, and 100 sequential edits.

\begin{table*}[ht] 
  \centering 
  \small 
  \resizebox{0.85\textwidth}{!}{%
    \begin{tabular}{l c c c c c c c c c}  
      \toprule 
      \textbf{Model} & \#Edits  
        & \textbf{STEM} & $\Delta$\textbf{STEM}  
        & \textbf{Humanities} & $\Delta$\textbf{Humanities} 
        & \textbf{Social Sci.} & $\Delta$\textbf{Social Sci.} 
        & \textbf{Other} & $\Delta$\textbf{Other} \\ 
      \midrule 
      \multirow{3}{*}{LLaMA-8B} 
        & 0 (Raw) & 58.53 & –     & 72.52 & –     & 77.13 & –     & 69.79 & –     \\ 
        & 50      & 56.17 & \textcolor{red}{–2.36} 
                   & 72.32 & \textcolor{red}{–0.20} 
                   & 74.96 & \textcolor{red}{–2.17} 
                   & 68.48 & \textcolor{red}{–1.31} \\ 
        & 100     & 54.57 & \textcolor{red}{–3.96} 
                   & 70.79 & \textcolor{red}{–1.73} 
                   & 72.62 & \textcolor{red}{–4.51} 
                   & 63.34 & \textcolor{red}{–6.45} \\ 
      \midrule 
      \multirow{3}{*}{LLaMA-3B} 
        & 0 (Raw) & 50.98 & –     & 65.14 & –     & 69.35 & –     & 62.47 & –     \\ 
        & 50      & 50.22 & \textcolor{red}{–0.76} 
                   & 64.19 & \textcolor{red}{–0.95} 
                   & 66.78 & \textcolor{red}{–2.57} 
                   & 60.22 & \textcolor{red}{–2.25} \\ 
        & 100     & 48.42 & \textcolor{red}{–2.56} 
                   & 61.89 & \textcolor{red}{–3.25} 
                   & 64.92 & \textcolor{red}{–4.43} 
                   & 56.61 & \textcolor{red}{–5.86} \\ 
      \midrule
      \multirow{3}{*}{Qwen-7B} 
        & 0 (Raw) & 69.01 & –     
                   & 77.51 & –     
                   & 83.14 & –     
                   & 73.99 & –     \\ 
        & 50      & 68.63 & \textcolor{red}{–0.38} 
                   & 77.23 & \textcolor{red}{–0.28} 
                   & 82.86 & \textcolor{red}{–0.28} 
                   & 63.57 & \textcolor{red}{–10.42} \\ 
        & 100     & 65.92 & \textcolor{red}{–3.09} 
                   & 76.41 & \textcolor{red}{–1.10} 
                   & 80.74 & \textcolor{red}{–2.40} 
                   & 57.48 & \textcolor{red}{–16.51} \\ 
      \bottomrule 
    \end{tabular}%
  } 
  \caption{General‐domain category accuracies and absolute drops $\Delta$ on MMLU across sequential medical edits} 
  \label{tab:rq4_common_domain_single} 
\end{table*}

After 100 sequential medical edits, all common‐domain categories exhibit performance declines, with “Other” (which includes business, health, and miscellaneous topics) suffering the largest drop (e.g., –6.45 pp for LLaMA-8B; –5.86 pp for LLaMA-3B; -16.51 for Qwen-7B). STEM accuracy decreases by 3.96 pp on 8B and 2.56 pp on 3B, while Social Sciences and Humanities show smaller but nontrivial declines. These results reinforce our RQ4‐F2 finding: medical‐focused updates not only impair specialized medical knowledge but also erode broader general capabilities. The consistent degradation across diverse categories underscores the challenge of maintaining out‐of‐domain performance during extensive sequential editing.

\subsection{Sequential Editing Impact with different Length of Knowledge}
\label{seq-sgr-len}

To investigate whether the introduction of longer rationales using SGR-Edit would lead to more rapid degradation of general capabilities on MMLU, we conducted experiments using LLaMA-3B as our base LLM. We employed AlphaEdit and MEMIT as sequential editing methods, testing their impact with different rationale token lengths (200-250, 150-200, and 100-150) on MMLU performance after 0, 20, and 50 sequential edits.

\begin{table*}[ht]
\centering
\resizebox{0.83\textwidth}{!}{%
\begin{tabular}{llcccccccc}
\toprule
\multicolumn{2}{l}{\textbf{Method}} & \textbf{\#Edits} & \textbf{Overall} & \textbf{$\Delta$Overall} & \textbf{Medical} & \textbf{$\Delta$Medical} & \textbf{Non-Medical} & \textbf{$\Delta$Non-Medical} \\
\midrule
\multicolumn{2}{l}{pre-edit} & 0 & 60.7 & -- & 63.7 & -- & 60.5 & -- \\
\midrule
\multirow{12}{*}{\textbf{AlphaEdit}} & SGR 200-250 & 50 & 59.2 & \textbf{-1.5} & 60.7 & \textbf{-3.0} & 59.3 & \textbf{-1.2} \\
& SGR 150-200 & 50 & 58.1 & \textbf{-2.6} & 60.5 & \textbf{-3.2} & 58.4 & \textbf{-2.1} \\
& SGR 100-150 & 50 & 59.6 & \textbf{-1.1} & 60.7 & \textbf{-3.0} & 59.8 & \textbf{-0.7} \\
\cline{2-9}
& GTA $<$10 & 50 & 60.0 & \textbf{-0.7} & 62.3 & \textbf{-1.4} & 59.9 & \textbf{-0.6} \\
& RE 50-100 & 50 & 59.3 & \textbf{-1.4} & 61.0 & \textbf{-2.7} & 59.4 & \textbf{-1.1} \\
& RE 100-50 & 50 & 59.0 & \textbf{-1.7} & 60.4 & \textbf{-3.3} & 59.0 & \textbf{-1.5} \\
\cline{2-9}
& SGR 200-250 & 20 & 60.3 & \textbf{-0.4} & 63.1 & \textbf{-0.6} & 60.1 & \textbf{-0.4} \\
& SGR 150-200 & 20 & 60.1 & \textbf{-0.6} & 62.1 & \textbf{-1.6} & 60.0 & \textbf{-0.5} \\
& SGR 100-150 & 20 & 60.6 & \textbf{-0.1} & 62.8 & \textbf{-0.9} & 60.4 & \textbf{-0.1} \\
\cline{2-9}
& GTA $<$10 & 20 & 60.4 & \textbf{-0.3} & 64.1 & \textbf{+0.4} & 60.8 & \textbf{+0.3} \\
& RE 50-100 & 20 & 60.6 & \textbf{-0.1} & 63.9 & \textbf{+0.2} & 60.3 & \textbf{-0.2} \\
& RE 100-50 & 20 & 60.6 & \textbf{-0.1} & 63.4 & \textbf{-0.3} & 60.3 & \textbf{-0.2} \\
\midrule
\multirow{12}{*}{\textbf{MEMIT}} & SGR 200-250 & 50 & 25.5 & \textbf{-35.2} & 22.8 & \textbf{-40.9} & 25.8 & \textbf{-34.7} \\
& SGR 150-200 & 50 & 25.7 & \textbf{-35.0} & 24.9 & \textbf{-38.8} & 26.2 & \textbf{-34.3} \\
& SGR 100-150 & 50 & 25.4 & \textbf{-35.3} & 22.8 & \textbf{-40.9} & 25.8 & \textbf{-34.7} \\
\cline{2-9}
& GTA $<$10 & 50 & 25.9 & \textbf{-34.8} & 26.3 & \textbf{-37.4} & 26.2 & \textbf{-34.3} \\
& RE 50-100 & 50 & 24.4 & \textbf{-36.3} & 26.2 & \textbf{-37.5} & 24.9 & \textbf{-35.6} \\
& RE 100-50 & 50 & 26.5 & \textbf{-34.2} & 25.5 & \textbf{-38.2} & 24.3 & \textbf{-36.2} \\
\cline{2-9}
& SGR 200-250 & 20 & 54.8 & \textbf{-5.9} & 42.3 & \textbf{-21.4} & 57.0 & \textbf{-3.5} \\
& SGR 150-200 & 20 & 51.4 & \textbf{-9.3} & 33.3 & \textbf{-30.4} & 54.7 & \textbf{-5.8} \\
& SGR 100-150 & 20 & 54.9 & \textbf{-5.8} & 45.7 & \textbf{-18.0} & 57.1 & \textbf{-3.4} \\
\cline{2-9}
& GTA $<$10 & 20 & 48.2 & \textbf{-12.5} & 33.0 & \textbf{-30.7} & 50.6 & \textbf{-9.9} \\
& RE 50-100 & 20 & 43.9 & \textbf{-16.8} & 27.4 & \textbf{-36.3} & 48.1 & \textbf{-12.4} \\
& RE 100-50 & 20 & 54.6 & \textbf{-6.1} & 40.7 & \textbf{-23.0} & 56.9 & \textbf{-3.6} \\
\bottomrule
\end{tabular}
}
\caption{Performance comparison of AlphaEdit and MEMIT for different knowledge lengths and number of edits on MMLU benchmark.}
\label{tab:seq-sgr-len-tab}
\end{table*}

The experimental results in Table \ref{tab:seq-sgr-len-tab} do not indicate that the introduction of longer rationales used for editing leads to a more rapid degradation of general capabilities on MMLU. In contrast, methods like MEMIT exhibit significantly higher degradation regardless of the rationale length, highlighting that \textbf{the degradation is more dependent on the editing method itself rather than solely on the rationale length.}

\subsection{Sequential Editing with ROME on External Domains}
\label{sec:rome-external}

To further verify the generalization of our findings in RQ4 (Section~\ref{sec-rq4}), we conducted extended experiments with ROME on LLaMA-3B. 
As shown in Table~\ref{tab:rome_ext}, similar to MEMIT, the model edited by ROME exhibits severe degradation in both medical and non-medical external domains as the number of edits increases. This contrasts sharply with the relative stability of AlphaEdit (see Figure~\ref{fig-seq-ext}), strongly supporting our conclusion that traditional KE methods fail to conserve external domain knowledge in sequential settings.

\begin{table}[ht]
\centering
\small
\resizebox{0.48\textwidth}{!}{%
\begin{tabular}{lcccc}
\toprule
\textbf{Method} & \textbf{\#Edits} & \textbf{Overall} & \textbf{Health} & \textbf{Non-Health} \\
\midrule
\multirow{3}{*}{\textbf{ROME}} 
& 0 & 60.7 & 63.7 & 60.5 \\
& 50 & 24.1 & 26.1 & 23.9 \\
& 100 & 22.9 & 23.9 & 23.0 \\
\bottomrule
\end{tabular}
}
\caption{Sequential editing impact of ROME on MMLU External Domains (LLaMA-3B).}
\label{tab:rome_ext}
\end{table}

\subsection{Sequential Stability Across Methods and Paradigms}
\label{sec:seq-paradigms}

In Section~\ref{sec-rq5}, we observed that MEMIT suffers from model collapse. To investigate whether this issue is method-specific or paradigm-dependent, we extend our analysis to include \textbf{AlphaEdit}, \textbf{ROME}, and \textbf{MEMIT} across all three paradigms (SGR-Edit, RE-Edit, GTA-Edit) on LLaMA-3B.

Table~\ref{tab:seq_paradigm_compare} presents the comprehensive results. We observe two key trends:
\begin{itemize}
    \item \textbf{Traditional Methods Collapse:} Both ROME and MEMIT exhibit consistent catastrophic collapse (accuracy dropping to $\sim$0.0\% on TriviaQA and GSM8K) after 50 edits, regardless of types of target knowledge. This indicates that the instability stems from the \textit{parameter-update mechanism} of these methods rather than the editing paradigm.
    \item \textbf{AlphaEdit Robustness:} In stark contrast, AlphaEdit maintains high stability across all paradigms. Even after 100 edits, SGR-Edit with AlphaEdit retains 62.2\% on GSM8K and 27.6\% on TriviaQA, significantly outperforming traditional approaches.
\end{itemize}

\begin{table*}[t]
\centering
\caption{Sequential editing performance comparison of \textbf{AlphaEdit}, \textbf{ROME}, and \textbf{MEMIT} across different editing paradigms on LLaMA-3B. \textcolor{red}{Red} indicates model collapse.}
\label{tab:seq_paradigm_compare}
\resizebox{0.95\textwidth}{!}{%
\begin{tabular}{ll|cccc|cccc|cccc} 
\toprule
\multirow{2}{*}{\textbf{Paradigm}} & \multirow{2}{*}{\textbf{\#Edits}} 
& \multicolumn{4}{c|}{\textbf{AlphaEdit}} 
& \multicolumn{4}{c|}{\textbf{ROME}} 
& \multicolumn{4}{c}{\textbf{MEMIT}} \\
\cmidrule(lr){3-6} \cmidrule(lr){7-10} \cmidrule(lr){11-14}
& & \textbf{Hella} & \textbf{Trivia} & \textbf{Truth} & \textbf{GSM} 
& \textbf{Hella} & \textbf{Trivia} & \textbf{Truth} & \textbf{GSM} 
& \textbf{Hella} & \textbf{Trivia} & \textbf{Truth} & \textbf{GSM} \\
\midrule
\multirow{3}{*}{SGR-Edit}
& 0   & 70.4 & 33.8 & 49.7 & 64.6 & 70.4 & 33.8 & 49.7 & 64.6 & 70.4 & 33.8 & 49.7 & 64.6 \\
& 50  & 70.7 & 31.6 & 49.9 & 64.1 & 24.7 & \textcolor{red}{0.0} & 51.1 & \textcolor{red}{0.0} & 26.2 & \textcolor{red}{0.0} & 51.5 & \textcolor{red}{0.0} \\
& 100 & 70.0 & 27.6 & 49.9 & 62.2 & 23.9 & \textcolor{red}{0.0} & 49.9 & \textcolor{red}{0.0} & 26.5 & \textcolor{red}{0.0} & 48.2 & \textcolor{red}{0.0} \\
\midrule
\multirow{3}{*}{RE-Edit}
& 0   & 70.4 & 33.8 & 49.7 & 64.6 & 70.4 & 33.8 & 49.7 & 64.6 & 70.4 & 33.8 & 49.7 & 64.6 \\
& 50  & 71.0 & 32.2 & 48.4 & 64.1 & 24.8 & \textcolor{red}{0.0} & 48.6 & \textcolor{red}{0.0} & 26.6 & \textcolor{red}{0.0} & 48.0 & \textcolor{red}{0.0} \\
& 100 & 70.8 & 29.3 & 48.7 & 62.8 & 25.7 & \textcolor{red}{0.0} & 48.3 & \textcolor{red}{0.0} & 26.4 & \textcolor{red}{0.0} & 47.9 & \textcolor{red}{0.0} \\
\midrule
\multirow{3}{*}{GTA-Edit}
& 0   & 70.4 & 33.8 & 49.7 & 64.6 & 70.4 & 33.8 & 49.7 & 64.6 & 70.4 & 33.8 & 49.7 & 64.6 \\
& 50  & 70.9 & 32.4 & 49.9 & 64.3 & 26.9 & \textcolor{red}{0.0} & 46.3 & \textcolor{red}{0.0} & 30.5 & \textcolor{red}{0.0} & 49.8 & \textcolor{red}{0.0} \\
& 100 & 70.9 & 31.2 & 48.9 & 63.5 & 26.2 & \textcolor{red}{0.0} & 48.7 & \textcolor{red}{0.0} & 26.2 & \textcolor{red}{0.0} & 47.2 & \textcolor{red}{0.0} \\
\bottomrule
\end{tabular}%
}
\end{table*}

\section{Case Analysis}
\label{app-cases}

\begin{figure*}[!htp]
  \centering
    \includegraphics[width=1\textwidth]{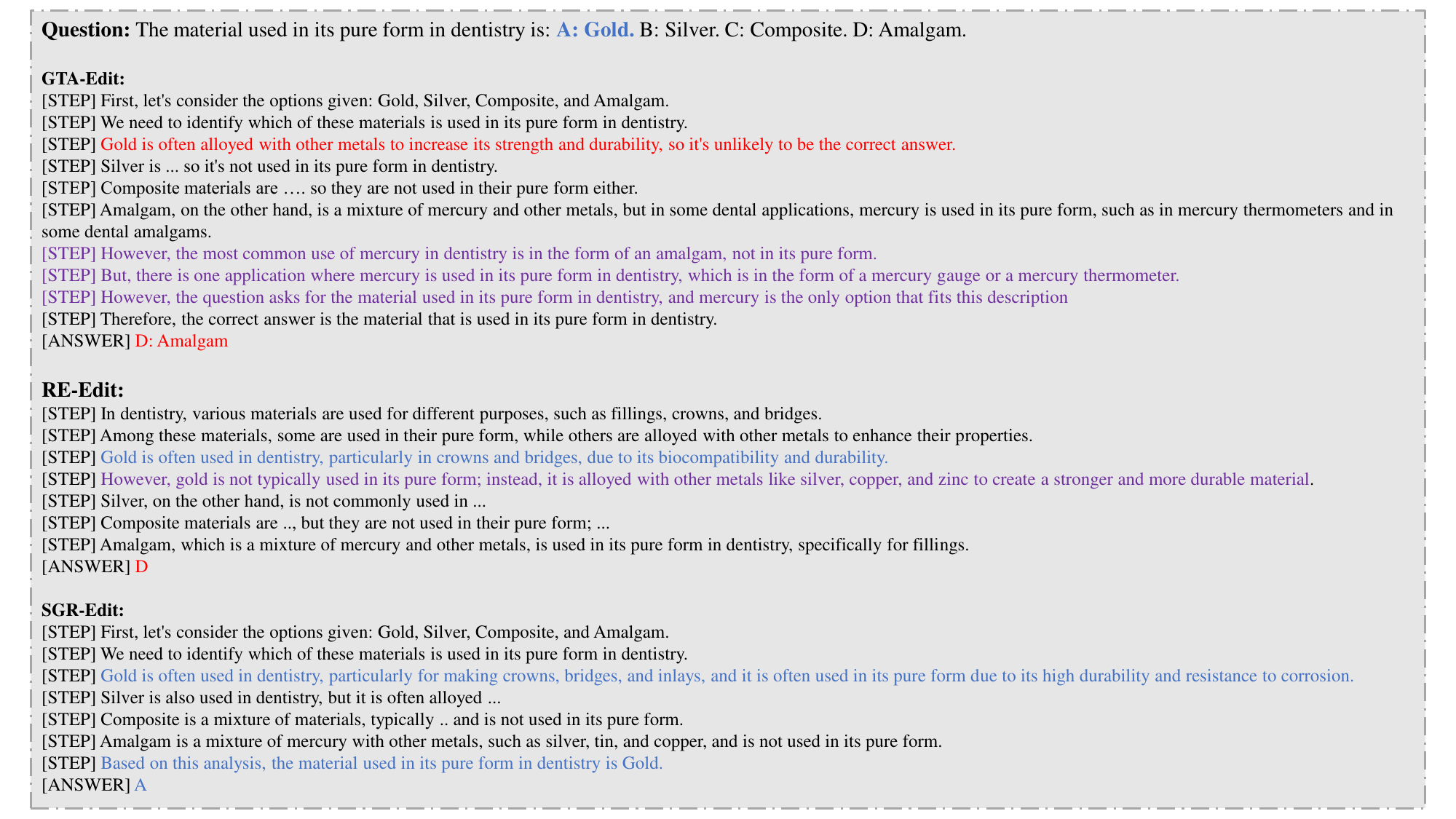}
  \caption{
    \textbf{Case: Medical knowledge editing performance across various editing paradigms.}
    The question asks which material is used in its pure form in dentistry. 
    \textcolor{blue}{Blue} indicates the correct post-edit answer, 
    \textcolor{red}{Red} denotes the pre-edit (incorrect) answer that the model memorized, 
    and \textcolor{lavenderblue}{Purple} highlights areas of reasoning confusion. 
    GTA-Edit fails to modify the model’s original reasoning path, causing it to revert to the pre-edit belief.
    RE-Edit introduces more context but still exhibits uncertainty.
    While SGR-Edit enables the LLM to internalize the knowledge and produce a clear, logically sound rationale.
  }
  \label{fig:edit_case_gold}
\end{figure*}

The case in Figure~\ref{fig:edit_case_gold}, using the MEMIT method to edit \textsc{LLaMA-8B} on MedMCQA\textsubscript{edit}, illustrates how different editing paradigms shape the reasoning behavior of post-edit LLM. This representative example highlights not only the final answer accuracy but also the depth of knowledge integration achieved by each paradigm. It reflects a broader pattern consistently observed across our benchmark: surface-level edits often fail to update the underlying reasoning logic, while rationale-driven editing enables models to produce more coherent and knowledge-grounded justifications.

Specifically, under \textbf{GTA-Edit}, although the correct answer \textcolor{blue}{“Gold”} was injected, the model struggles to abandon its prior belief \textcolor{red}{“Amalgam”}, which was likely encoded during pretraining. Instead of incorporating the new knowledge, the model retains previous reasoning patterns (\textcolor{red}{red} segments) and ends up with contradictory justifications (\textcolor{lavenderblue}{purple} segments), reflecting surface-level memorization rather than meaningful conceptual update.

\textbf{RE-Edit} provides more context about material usage but still fails to fully overwrite the original logic. The rationale demonstrates partial knowledge integration (\textcolor{blue}{blue} segments), yet the model remains hesitant (\textcolor{lavenderblue}{purple} segments) and eventually falls back on incorrect elimination logic.

In contrast, \textbf{SGR-Edit} delivers a clear and coherent rationale. The model identifies that gold is used in its pure form due to its corrosion resistance and biocompatibility (\textcolor{blue}{blue} segments), precisely the kind of causal justification absent in GTA/RE edits. This shows that editing with a reasoning-rich target enables the LLM to internalize and apply the knowledge reliably. Such examples highlight the importance of reasoning-grounded editing for domains like medicine, where interpretability and trustworthiness are critical.

\end{document}